\documentclass{article}

\usepackage{arxiv}
\usepackage[utf8]{inputenc} 
\usepackage[T1]{fontenc}    
\usepackage{hyperref}       
\usepackage{url}            
\usepackage{booktabs}       
\usepackage{amsfonts}       
\usepackage{nicefrac}       
\usepackage{microtype}      
\usepackage{lipsum}
\usepackage{graphicx}
\graphicspath{ {./images/} }

\usepackage{amsmath,amsfonts,bm}

\def\eqref#1{equation~\ref{#1}}

\def\1{\bm{1}}

\DeclareMathAlphabet{\mathsfit}{\encodingdefault}{\sfdefault}{m}{sl}
\SetMathAlphabet{\mathsfit}{bold}{\encodingdefault}{\sfdefault}{bx}{n}

\usepackage{natbib}
\usepackage{graphicx}
\usepackage{booktabs}
\usepackage{multirow}
\usepackage{amsmath,amssymb,amsfonts}
\usepackage{amsthm}
\usepackage{mathrsfs}
\usepackage[title]{appendix}
\usepackage{xcolor}
\usepackage{textcomp}
\usepackage{manyfoot}
\usepackage{algorithm}
\usepackage{algorithmicx}
\usepackage{algpseudocode}
\usepackage{listings}
\usepackage{float}
\usepackage{tikz}
\usetikzlibrary{positioning, arrows.meta}
\usepackage{adjustbox,lipsum}
\usepackage{silence}
\ErrorsOff*

\newtheorem{definition}{Definition}%

\title{CuBAS: Information Geometric Curvature-Based Adaptive Sampling for Supervised Classification}

\author{
 Alexandre Luis Magalh\~aes Levada\\
  Federal University of S\~ao Carlos\\
  13565-905, S\~ao Carlos-SP, Brazil\\
  \texttt{alexandre.levada@ufscar.br} \\
}

\begin{document}
\maketitle
\begin{abstract}
The informativeness of a training set is as consequential as its size, yet most sampling strategies remain agnostic to the intrinsic geometry of the data distribution. We address this gap by introducing CuBAS (Curvature-Based Adaptive Sampling), an information-geometric framework for adaptive data selection in supervised classification, grounded in the q-state Potts Markov random field (MRF) model. The central insight is that a labeled dataset can be viewed as a statistical manifold, on which local curvature, estimated via the ratio of second- to first-order observed Fisher information, faithfully encodes the geometric complexity of the underlying data distribution. Concretely, we construct a k-nearest-neighbor graph over the labeled data and derive a closed-form curvature score at each vertex from the Potts sufficient statistics, avoiding costly eigendecompositions or kernel density estimates. This curvature signal naturally partitions the graph into two complementary regimes: low-curvature regions, corresponding to smooth, homogeneous clusters that are efficiently represented by a small number of prototypical samples, and high-curvature regions, concentrated around decision boundaries and topologically complex structures that are disproportionately informative for classification. By selecting nodes from both regimes in a principled, geometry-aware manner, CuBAS constructs compact yet maximally informative training subsets. Extensive empirical evaluation across 30 benchmark datasets, spanning tabular, image, and biological domains, demonstrates consistent and statistically significant improvements in classification accuracy over random sampling and uncertainty-based baselines, across a wide range of labeling budgets and classifier architectures. Our method is computationally efficient (linear in the number of edges of the k-NN graph), theoretically grounded in the differential geometry of statistical manifolds, and directly interpretable in terms of the local shape operator of the data manifold. CuBAS thus offers a principled, scalable, and geometry-aware alternative to heuristic sampling for supervised learning.
\end{abstract}

\section{Introduction}
\label{sec:introduction}

The remarkable success of modern supervised learning has been driven not only by increasingly sophisticated learning algorithms, but also by the availability of large annotated datasets. Nevertheless, the effectiveness of a classifier depends far more on the \emph{quality} and representativeness of its training samples than on their sheer quantity. Large datasets often contain substantial redundancy, noisy observations, class imbalance, and numerous samples that contribute little to defining the underlying decision boundaries. This observation naturally motivates a fundamental question: \emph{how can we identify the most informative samples in a labeled dataset for classification?}

Addressing this question is becoming increasingly important as datasets continue to grow in size and complexity. Although considerable effort has been devoted to designing more expressive classifiers, comparatively less attention has been paid to principled mechanisms for selecting informative training samples in a way that reflects the intrinsic geometric organization of the data \cite{Wilson1972}. Traditional sampling and instance reduction techniques, including random subsampling, prototype selection, and instance weighting, are typically heuristic and operate primarily in the input space, making limited use of the statistical relationships among neighboring samples \cite{Garcia2012}. As a consequence, they often fail to distinguish truly informative boundary samples from redundant observations located in homogeneous regions.

Recent advances in graph-based learning have demonstrated that representing datasets as graphs provides a powerful framework for capturing local and global relationships among samples \cite{Zhu2003,Belkin2006}. Graph representations naturally encode neighborhood interactions and have become fundamental tools in manifold learning, semi-supervised learning, graph neural networks, and spectral methods. From this perspective, the informative content of a sample is not solely determined by its individual attributes, but also by its structural role within the graph topology. In particular, vertices located near class transitions or heterogeneous neighborhoods frequently carry substantially more discriminative information than vertices embedded in homogeneous regions. Despite these advances, existing graph-based sampling strategies rarely exploit the differential geometry induced by statistical interactions on the graph, relying instead on fixed similarity measures or purely combinatorial criteria.

Geometric Machine Learning has recently emerged as a unifying paradigm that exploits the geometric structure of data to design learning algorithms that are more robust, interpretable, and statistically efficient \cite{bronstein2017geometric}. Rather than treating observations as isolated points in a Euclidean feature space, geometric learning represents data through manifolds, graphs, or other non-Euclidean domains, allowing algorithms to explicitly leverage neighborhood relationships, topology, and curvature \cite{Papillon2025}. This perspective has driven significant advances in manifold learning, graph neural networks, geometric deep learning, and information geometry, demonstrating that incorporating geometric priors often leads to improved generalization and more faithful representations of complex data \cite{GML}. In contrast to most existing geometric learning methods, which primarily focus on learning feature representations or graph embeddings, the proposed approach employs geometry to address a different but equally fundamental problem: identifying the most informative training samples. By modeling labeled $k$-NN graphs as statistical manifolds induced by a Potts Markov random field and quantifying local geometric complexity through Fisher-information-based curvature, CuBAS transforms adaptive sampling into a geometric inference problem. Consequently, the proposed method extends the scope of geometric machine learning beyond representation learning, showing that differential-geometric quantities can also provide principled criteria for sample selection, redundancy reduction, and classifier training.

Information geometry offers a natural mathematical framework to bridge this gap. By viewing a statistical model as a Riemannian manifold endowed with the Fisher information metric, it becomes possible to characterize the local geometry of probability distributions through curvature and other differential-geometric quantities \cite{Amari2016,Nielsen}. Curvature provides a quantitative measure of how rapidly the statistical model changes locally, making it a natural descriptor of sample informativeness. Surprisingly, despite its strong theoretical foundations, information geometry has received relatively little attention as a mechanism for adaptive sampling and data reduction in supervised learning.

The problem of selecting informative samples has been extensively investigated under several paradigms, including active learning, curriculum learning, and instance or prototype selection. Active learning seeks to minimize labeling effort by querying informative unlabeled samples according to uncertainty, margin, or expected model change criteria \cite{Tong2002,settles2010active}. While highly effective in interactive settings, these methods are inherently classifier-dependent and primarily designed for scenarios in which labels are unavailable. Curriculum learning adopts a complementary perspective by progressively presenting training samples according to a predefined notion of difficulty \cite{Bengio2009}. Although successful in many deep learning applications, curriculum strategies generally rely on heuristic measures of sample complexity and lack a rigorous geometric interpretation. Instance and prototype selection methods attempt to reduce dataset redundancy through distance-based, density-based, or boundary-preserving criteria \cite{Wilson1972,Garcia2012,Liu2002,Aggarwal2014,Lopez2010,Aida2019}. However, these approaches typically operate in the original feature space using fixed metrics and do not explicitly account for the statistical geometry induced by class interactions. Likewise, graph-based approaches incorporate neighborhood information but seldom provide a principled characterization of local statistical complexity or boundary sharpness \cite{Zhou2003,Belkin2006,Zhang2022}.

This paper introduces a fundamentally different perspective. Rather than measuring sample importance through classifier uncertainty, local density, or heuristic boundary estimates, we define informativeness as an intrinsic geometric property of a statistical manifold induced by a labeled graph. Specifically, we model the graph using a $q$-state Potts Markov random field, whose local interactions define a probabilistic manifold parameterized by the inverse temperature. Within this framework, first- and second-order Fisher information naturally induce the metric and curvature of the manifold, enabling the construction of a local shape operator that quantifies the geometric complexity surrounding each vertex. Consequently, sample selection becomes a problem of identifying regions of high statistical curvature, where class interactions are strongest and the information content is maximal.

Building upon this geometric formulation, we propose the \emph{Curvature-Based Adaptive Sampling} (CuBAS) algorithm, a graph-based, model-agnostic sampling strategy that decomposes a labeled dataset into complementary subsets according to their local information curvature. Samples exhibiting high curvature are predominantly located near decision boundaries and heterogeneous neighborhoods, whereas low-curvature samples correspond to smooth, homogeneous regions of the data manifold. Unlike existing sampling techniques, CuBAS derives its selection criterion directly from the geometry of the underlying statistical model, providing a unified framework that simultaneously captures boundary awareness, redundancy reduction, and noise filtering.

The main contributions of this work are summarized as follows:

\begin{itemize}
	\item We introduce a novel graph-based information-geometric framework that models labeled datasets as statistical manifolds induced by a $q$-state Potts Markov random field, providing a rigorous connection between graph topology and differential geometry.
	
	\item We derive closed-form expressions for the first- and second-order Fisher information of the Potts model and show how they induce a local shape operator that quantifies the statistical curvature associated with each graph vertex.
	
	\item We propose CuBAS, a computationally efficient curvature-driven adaptive sampling algorithm that identifies informative samples through local geometric analysis rather than heuristic or classifier-dependent criteria.
	
	\item We demonstrate experimentally, using a diverse collection of benchmark datasets and multiple classification algorithms, that CuBAS consistently produces graph decompositions with stronger community structure and yields higher classification accuracy than conventional random sampling.
\end{itemize}

The remainder of this paper is organized as follows. Section~2 reviews the $q$-state Potts Markov random field and the maximum pseudo-likelihood procedure used for parameter estimation. Section~3 develops the proposed information-geometric framework, derives the Fisher information tensors, and introduces the curvature-based adaptive sampling methodology. Section~4 presents an extensive experimental evaluation on benchmark datasets and analyzes both the geometric and classification performance of the proposed approach. Finally, Section~5 concludes the paper and discusses promising directions for future research.

\section{Theoretical Background}
\label{sec:theory}
This section introduces the theoretical foundations underlying the proposed \emph{Curvature-Based Adaptive Sampling} (CuBAS) framework. Our approach builds upon three complementary mathematical components that establish the connection between graph-based statistical modeling and differential geometry. First, we review the isotropic $q$-state Potts Markov Random Field (MRF), which provides a probabilistic model for describing the interactions among neighboring labels in a $k$-nearest neighbor graph. Next, we present the maximum pseudo-likelihood (MPL) estimation procedure used to efficiently estimate the inverse temperature parameter governing the Potts model. Finally, we summarize the differential geometric concepts that form the basis of our information-geometric analysis, including the metric tensor, the curvature tensor, and the shape operator. These concepts provide the mathematical tools required to characterize the local geometry of the statistical manifold induced by the Potts model and ultimately define the curvature-based criterion employed by CuBAS to identify informative samples. The proposed framework emerges from the integration of probabilistic graphical models and information geometry. The Potts MRF captures local statistical dependencies on labeled graphs, maximum pseudo-likelihood provides an efficient parameter estimation procedure, and differential geometry supplies the geometric descriptors that quantify the statistical complexity of each neighborhood.

\subsection{The isotropic Potts Markov random field}

The \emph{Potts model} is a discrete \emph{Markov random field} (MRF) in which each random variable assumes one of a finite number of states \( C = \{1,2,\ldots,q\} \). Originally introduced to study interacting spin systems, the Potts model has since evolved into a general-purpose probabilistic framework for modeling contextual dependence among discrete variables. Its defining characteristic is the explicit encoding of pairwise interactions that favor state agreement between neighboring sites, allowing complex global structures to emerge from simple local rules. As a result, the Potts model naturally captures phenomena such as clustering, spatial coherence, and phase transitions, which are central to many problems involving structured data. Due to this expressive yet interpretable formulation, the model has been widely adopted across mathematics \cite{PottsMath2,PottsMath3,PottsMath4}, statistical physics \cite{PottsPhys1,PottsPhys2,PottsPhys3}, computational biology \cite{PottsBiol1,PottsBiol2,PottsBiol3}, and pattern recognition and computer vision \cite{PottsCS1,PottsCs2}, where it serves as a canonical model for interaction-driven dynamics on graphs and lattices.

From a probabilistic standpoint, the Hammersley-Clifford theorem \cite{Hammersley} guarantees that the Potts model admits an equivalent representation either as a global Gibbs distribution or as a collection of local conditional distributions defining a Markov random field. In this work, we adopt the local conditional formulation, which is particularly well suited for graph-based learning settings. This choice enables direct evaluation of node-level conditional probabilities given their neighborhoods, while avoiding the intractable partition function associated with the global Gibbs measure. Let \(\eta_i\) denote the neighborhood of node \(i\). The local conditional distribution of the isotropic pairwise \(q\)-state Potts model is then given by \cite{LCDF}

\begin{equation}
	p\left( x_i = m \mid \eta_i, \beta \right)
	=
	\frac{\exp\!\left( \beta\, U_i(m) \right)}
	{\sum_{\ell=1}^{q} \exp\!\left( \beta\, U_i(\ell) \right)},
\end{equation}

where \(U_i(m)\) denotes the number of neighbors of node \(i\) whose labels equal \(m\), \(q\) is the total number of states, and \(\beta\) is the inverse temperature parameter controlling the strength of contextual dependence. The parameter \(\beta\) governs the smoothness of the resulting label configurations: larger values promote strong agreement among neighboring nodes, leading to homogeneous regions and well-defined clusters, whereas \(\beta = 0\) yields a degenerate case of statistical independence in which all labels are equiprobable and contextual structure vanishes. This explicit control over interaction strength makes the Potts model particularly attractive for studying how local label interactions shape the geometry and structure of labeled graphs, a property that is central to the curvature-based analysis developed in this work.

\subsection{Estimation of the inverse temperature parameter}

Estimating coupling parameters in Markov random field (MRF) models is a well-known challenge due to the presence of the partition function in the joint Gibbs distribution, which makes maximum likelihood (ML) estimation computationally prohibitive for large-scale systems \cite{BesagMPL}. This difficulty is particularly acute in graph-based settings, where the number of configurations grows exponentially with the number of nodes. A widely adopted and theoretically sound alternative is the \emph{maximum pseudo-likelihood} (MPL) method \cite{BesagMPL}, which approximates the global likelihood by a product of local conditional probabilities defined over node neighborhoods. By replacing the joint distribution with these local factors, the MPL approach eliminates the need to compute the partition function, thereby yielding a tractable optimization problem while retaining essential statistical dependencies.

Beyond its computational appeal, the MPL estimator enjoys strong asymptotic properties, including consistency and asymptotic normality under mild regularity conditions \cite{MPL}. These guarantees make pseudo-likelihood-based estimation a robust and interpretable choice for parameter inference in structured probabilistic models, including Potts and Ising-type MRFs. For the isotropic Potts model considered in this work, the pseudo-likelihood function is defined as
\begin{equation}
	PL(\beta)
	=
	\prod_{i=1}^{n} p\left( x_i \mid \eta_i, \beta \right)
	=
	\prod_{i=1}^{n}
	\left[
	\frac{\exp\!\left( \beta\, U_i(x_i) \right)}
	{\sum_{\ell=1}^{q} \exp\!\left( \beta\, U_i(\ell) \right)}
	\right],
\end{equation}
where \(n\) denotes the number of nodes in the graph, \(U_i(x_i)\) is the number of neighbors of node \(i\) sharing the same label as \(x_i\), \(q\) is the number of possible states, and \(\beta\) is the inverse temperature parameter controlling the strength of spatial coupling.

In practice, it is more convenient to maximize the log-pseudo-likelihood, which transforms the product of local terms into a summation and improves numerical stability. The log-pseudo-likelihood is given by
\begin{equation}
	\log PL(\beta)
	=
	\sum_{i=1}^{n}
	\left[
	\beta\, U_i(x_i)
	-
	\log \left(
	\sum_{\ell=1}^{q}
	\exp\!\left( \beta\, U_i(\ell) \right)
	\right)
	\right].
\end{equation}
The first term measures the contribution of the observed configuration, while the second term accounts for normalization over all possible neighborhood label assignments. Differentiating this expression with respect to \(\beta\) and setting the derivative to zero yields the MPL estimating equation
\begin{equation}
	\sum_{i=1}^{n} U_i(x_i)
	-
	\sum_{i=1}^{n}
	\left[
	\frac{
		\sum_{\ell=1}^{q}
		U_i(\ell)\exp\!\left( \beta\, U_i(\ell) \right)
	}{
		\sum_{\ell=1}^{q}
		\exp\!\left( \beta\, U_i(\ell) \right)
	}
	\right]
	= 0.
\end{equation}
Looking carefully to the pseudo-likeilhood equation, we see that the first term is the observed energy of the system, whereas the second one represents the expected energy under the model conditioned on the current value of \(\beta\). Intuitively, the MPL estimate of the inverse temperature parameter is obtained when the empirical and model-predicted neighborhood agreement are in balance.

The resulting nonlinear equation does not admit a closed-form solution and must be solved numerically. In this work, we employ the secant method \cite{Burden}, a derivative-free root-finding algorithm that offers a favorable trade-off between computational simplicity and convergence speed. Let \(f(\beta)\) denote the left-hand side of the MPL estimating equation. Given two initial estimates \(\beta_0\) and \(\beta_1\), the secant method generates a sequence \(\{\beta_k\}\) according to
\begin{equation}
	\beta_{k+1}
	=
	\beta_k
	-
	f(\beta_k)\,
	\frac{\beta_k - \beta_{k-1}}{f(\beta_k) - f(\beta_{k-1})}.
\end{equation}
At each iteration, the derivative of \(f\) is approximated by a finite difference between successive iterates, and the next estimate is obtained from the intersection of the corresponding secant line with the horizontal axis. Under standard smoothness assumptions, the secant method achieves superlinear convergence and typically outperforms bisection-based approaches, while avoiding the explicit computation of derivatives required by Newton-Raphson methods. These properties make it particularly suitable for efficient estimation of the inverse temperature parameter in large-scale Potts models.

\paragraph{Computational complexity and scalability.}
From a computational perspective, the MPL-based estimation of the inverse temperature parameter scales linearly with the number of nodes in the graph. For each evaluation of the objective function \(f(\beta)\), the dominant cost arises from computing the neighborhood agreement terms \(U_i(\ell)\) for all nodes \(i\) and labels \(\ell \in \{1,\ldots,q\}\), resulting in a complexity of \(O(n\, q\, \bar{d})\), where \(n\) is the number of nodes and \(\bar{d}\) denotes the average node degree. In typical pattern recognition applications, both \(q\) and \(\bar{d}\) are small constants, making the overall cost effectively linear in \(n\). Moreover, the secant method converges in a small number of iterations, as only a single scalar parameter is estimated. Consequently, the entire parameter estimation procedure introduces negligible overhead relative to graph construction or classifier training. This favorable scaling behavior allows the proposed framework to be applied to moderately large datasets and dense graph representations without compromising computational efficiency.

\subsection{Differential geometry basics}

Differential geometry provides a rigorous mathematical framework for analyzing the local and global structure of smooth manifolds. In recent years, geometric concepts have become increasingly important in machine learning, where high-dimensional datasets are often modeled as samples lying on low-dimensional manifolds embedded in ambient spaces. Within this perspective, geometric quantities such as distances, metric tensors, and curvature provide principled measures of local complexity, enabling the characterization of regions with distinct statistical and structural properties \citep{spivak1999comprehensive, ONeill2006, do_carmo_differential_2016,	tu2017differential, needham2021visual}.

The proposed CuBAS framework adopts this viewpoint by interpreting the parameter space of the isotropic $q$-state Potts Markov random field as a statistical manifold endowed with a Riemannian structure induced by Fisher information. Rather than studying the geometry of the feature space directly, we analyze the geometry of the probabilistic model governing the labeled $k$-NN graph. In this setting, curvature becomes an intrinsic descriptor of how rapidly the statistical model changes with respect to its parameters, providing a natural measure of sample informativeness.

This section briefly reviews the differential-geometric concepts required throughout the remainder of the paper. In particular, we introduce the first and second fundamental forms, the metric tensor, the curvature tensor, and the shape operator, which constitute the mathematical foundation of the proposed curvature-based adaptive sampling strategy.

\subsubsection{Metric Tensor and First Fundamental Form}

Let $M$ be a smooth manifold parameterized by local coordinates
$\theta=(\theta_1,\ldots,\theta_m)$ through a differentiable mapping

\begin{equation}
	\mathbf{X}(\theta):\mathbb{R}^{m}\rightarrow\mathbb{R}^{n}.
\end{equation}

The first fundamental form defines the intrinsic geometry of the manifold by measuring infinitesimal distances between neighboring points. It is represented by the metric tensor

\begin{equation}
	g_{ij}(\theta)
	=
	\left\langle
	\frac{\partial \mathbf{X}}{\partial\theta_i},
	\frac{\partial \mathbf{X}}{\partial\theta_j}
	\right\rangle,
\end{equation}

where $\langle\cdot,\cdot\rangle$ denotes the Euclidean inner product. The corresponding matrix

\[
\mathbf{G}(\theta)=\left[g_{ij}(\theta)\right]
\]

encodes all intrinsic geometric information, including distances, angles, and local volumes.

In information geometry, the role of the metric tensor is naturally played by the Fisher Information Matrix (FIM), which equips the statistical manifold with a Riemannian structure \cite{Amari2016,Nielsen}. Consequently, nearby probability distributions become points on a curved manifold, whose geometry reflects the sensitivity of the underlying probabilistic model to parameter perturbations.

\subsubsection{Second Fundamental Form}

While the metric tensor characterizes intrinsic geometry, it does not describe how the manifold bends within its embedding space. This extrinsic behavior is captured by the second fundamental form.

Let $\mathbf{N}(\theta)$ denote the unit normal vector field of the manifold. The coefficients of the second fundamental form are defined as

\begin{equation}
	b_{ij}(\theta)
	=
	\left\langle
	\frac{\partial^2\mathbf{X}}
	{\partial\theta_i\partial\theta_j},
	\mathbf{N}
	\right\rangle,
\end{equation}

forming the symmetric matrix

\[
\mathbf{B}(\theta)=\left[b_{ij}(\theta)\right].
\]

Whereas the first fundamental form measures infinitesimal distances along the manifold, the second fundamental form quantifies how rapidly the tangent plane changes from one point to another. Therefore, it provides direct information about local bending and curvature.

Within the proposed framework, this distinction has an important statistical interpretation. The first-order Fisher information measures the local sensitivity of the likelihood function, whereas the second-order Fisher information describes how this sensitivity itself varies over the parameter space. Their combination ultimately determines the local geometric complexity of the statistical manifold associated with the Potts model.

\subsubsection{Shape Operator}

The shape operator establishes the relationship between the intrinsic and extrinsic geometries of a manifold by combining the first and second fundamental forms into a single linear operator. It is one of the central objects in classical differential geometry because its spectral properties completely characterize local curvature.

\begin{definition}[Shape Operator]
	Let $M$ be a regular manifold endowed with the first and second fundamental forms,
	$\mathbf{G}(\theta)$ and $\mathbf{B}(\theta)$, respectively. The shape operator is defined as
	
	\begin{equation}
		\mathbf{S}(\theta)
		=
		-\mathbf{B}(\theta)\mathbf{G}(\theta)^{-1}.
	\end{equation}
	
\end{definition}

Geometrically, the shape operator measures how the unit normal vector changes when moving along tangent directions on the manifold. Equivalently, it quantifies the local rate at which the manifold bends in each direction. The eigenvectors of $\mathbf{S}$ define the principal curvature directions, while its eigenvalues correspond to the principal curvatures. Consequently, several classical curvature measures are obtained directly from $\mathbf{S}$, including the Gaussian curvature (determinant), the mean curvature (half the trace), and the principal curvatures (eigenvalues) \cite{Manfredo}.

\begin{figure}
	\centering
	\includegraphics[scale=0.5]{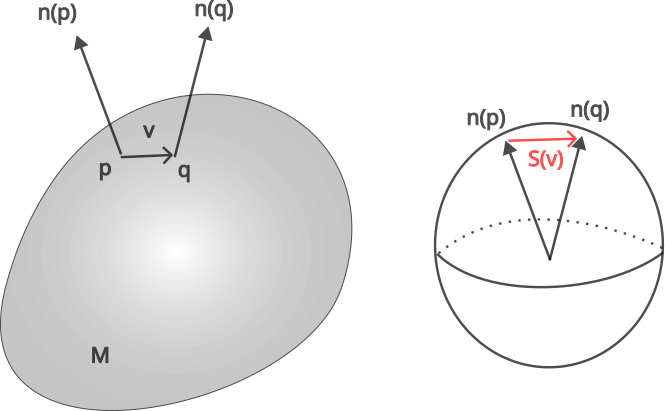}
	\caption{Geometric interpretation of the shape operator. As one moves along a tangent direction $\mathbf{v}$, the shape operator measures the directional variation of the unit normal vector, providing a quantitative description of the local bending of the manifold.}
	\label{fig:shape}
\end{figure}

Figure~\ref{fig:shape} illustrates the geometric interpretation of the shape operator. As one moves along a tangent direction, the operator measures the directional variation of the unit normal vector, thereby quantifying the local bending of the surface.

The proposed CuBAS method exploits this geometric interpretation in the context of statistical learning. Rather than computing curvature on a geometric surface embedded in Euclidean space, we evaluate the shape operator on the statistical manifold induced by the Potts Markov random field. Since the isotropic Potts model is parameterized solely by the inverse temperature parameter $\beta$, the statistical manifold is one-dimensional, causing the shape operator to reduce to a scalar quantity. This scalar curvature serves as a direct measure of local statistical complexity, allowing samples to be ranked according to the geometric information encoded in their neighborhood configurations. As demonstrated in the following sections, this curvature measure naturally separates homogeneous regions from decision boundaries, providing the theoretical foundation for the proposed curvature-based adaptive sampling strategy.

\section{CuBAS: curvature-based adaptive sampling}
\label{sec:method}

The proposed CuBAS framework is grounded in the principles of information geometry, where statistical models are interpreted as smooth manifolds endowed with a Riemannian structure induced by the Fisher information. In this setting, the notion of sample informativeness is no longer purely heuristic or model-dependent, but instead arises from intrinsic geometric properties of the underlying probability distributions. Intuitively, regions of the statistical manifold exhibiting high curvature correspond to parameter configurations where small perturbations induce large changes in the likelihood, signaling increased sensitivity, reduced redundancy, and greater discriminative potential. Conversely, flat regions indicate statistical homogeneity, where samples are largely interchangeable and contribute limited additional information. This geometric viewpoint provides a principled foundation for adaptive sampling, directly linking information content, boundary complexity, and learning efficiency.

Formally, the Fisher information quantifies the sensitivity of the likelihood function to infinitesimal perturbations of the model parameters. From a geometric perspective, this sensitivity induces a Riemannian metric on the statistical manifold associated with a parametric family of probability distributions. Within this framework, first-order Fisher information characterizes the local metric structure, determining how distances between neighboring distributions are measured, while second-order Fisher information captures curvature effects, reflecting how this metric varies across the manifold. These curvature effects play a central role in distinguishing statistically homogeneous regions from structurally complex regions, such as class boundaries, which are precisely the regions of interest for adaptive sampling.

\begin{definition}[First-order Fisher information]
	Let \(p(X;\theta)\) be a probability density (or mass) function parameterized by \(\theta \in \Theta \subset \mathbb{R}\).
	The \emph{first-order Fisher information} measures the expected sensitivity of the log-likelihood function to infinitesimal changes in \(\theta\) and is defined as \cite{Fisher}
	\begin{equation}
		\mathcal{I}(\theta)
		=
		\mathbb{E}\!\left[
		\left(
		\frac{\partial}{\partial \theta}
		\log p(X;\theta)
		\right)^{2}
		\right].
	\end{equation}
\end{definition}

\begin{definition}[Second-order Fisher information]
	Let \(p(X;\theta)\) be a probability density (or mass) function parameterized by \(\theta \in \Theta\).
	The \emph{second-order Fisher information} captures the curvature of the log-likelihood function with respect to the parameter \(\theta\) and is given by \cite{Fisher}
	\begin{equation}
		\mathcal{II}(\theta)
		=
		-\,\mathbb{E}\!\left[
		\frac{\partial^{2}}{\partial \theta^{2}}
		\log p(X;\theta)
		\right].
	\end{equation}
\end{definition}

From the standpoint of information geometry, the first-order Fisher information defines the metric tensor of the statistical manifold, endowing the parameter space with a notion of local distance and volume. The second-order Fisher information, in turn, characterizes how this metric varies locally, encoding curvature properties that reflect higher-order statistical interactions (curvature tensor). Together, these quantities enable a differential-geometric analysis of statistical models, in which notions such as geodesics, curvature, and shape operators become well defined. In the context of CuBAS, this geometric structure is exploited to identify regions of high information curvature on labeled graphs, providing the theoretical basis for distinguishing low-information, redundant samples from high-information samples concentrated near decision boundaries. The explicit derivation of the Fisher information tensors for the Potts model and their role in curvature-based adaptive sampling are detailed in the subsequent sections.

\subsection{Fisher information in the q-state Potts model}

The Fisher information associated with the \textit{$q$}-state Potts model provides a principled and geometrically grounded mechanism for characterizing statistical dependencies in labeled Markov random fields. From the perspective of information geometry, the Potts model induces a parametric statistical manifold whose geometry is governed by the sensitivity of the Gibbs distribution to perturbations in its coupling parameter, $\beta$. In this setting, the Fisher information plays a dual role: its first-order term defines a Riemannian metric on the parameter space, while its second-order term captures intrinsic curvature properties of the induced manifold. Together, these quantities encode how local label configurations respond to variations in interaction strength, thereby offering a natural notion of informational saliency within the field.

Let $p(x_i \mid \eta_i, \beta)$ denote the local conditional distribution of the isotropic pairwise Potts model, where $x_i$ is the label at node $i$ and $\eta_i$ denotes its neighborhood. The (scalar) Fisher information with respect to $\beta$ is defined as
\begin{equation}
	\mathcal{I}(\beta)
	=
	\mathbb{E}\!\left[
	\left(
	\frac{\partial}{\partial \beta}
	\log p(x_i \mid \eta_i, \beta)
	\right)^2
	\right],
\end{equation}
while the second-order Fisher information is given by
\begin{equation}
	\mathcal{II}(\beta)
	=
	-
	\mathbb{E}\!\left[
	\frac{\partial^2}{\partial \beta^2}
	\log p(x_i \mid \eta_i, \beta)
	\right].
\end{equation}
Under standard regularity conditions, both quantities can be interpreted as local descriptors of the curvature of the log-likelihood surface induced by the Potts model.

In practice, the expectations above are intractable due to the combinatorial nature of the Potts field. Following the law of large numbers \cite{ObsFisher}, we approximate these expectations empirically using observed samples drawn from the labeled graph. Let $\hat{\beta}$ denote the maximum pseudo-likelihood estimate of the coupling parameter. The first-order Fisher information can then be approximated as
\begin{equation}
	\mathcal{I}(\beta)
	\;\approx\;
	\Phi(\beta)
	=
	\frac{1}{n}
	\sum_{i=1}^{n}
	\left[
	\frac{\partial}{\partial \beta}
	\log p(x_i \mid \eta_i, \hat{\beta})
	\right]^2,
\end{equation}
while the second-order Fisher information is estimated by
\begin{equation}
	\mathcal{II}(\beta)
	\;\approx\;
	\Psi(\beta)
	=
	-
	\frac{1}{n}
	\sum_{i=1}^{n}
	\frac{\partial^2}{\partial \beta^2}
	\log p(x_i \mid \eta_i, \hat{\beta}).
\end{equation}

A key advantage of this formulation is that the global Fisher information naturally decomposes into local contributions associated with individual nodes of the labeled $k$-NN graph. Specifically, we define the node-wise first- and second-order Fisher information as
\begin{align}
	\Phi_i(\beta)
	&=
	\left[
	\frac{\partial}{\partial \beta}
	\log p(x_i \mid \eta_i, \hat{\beta})
	\right]^2,
	\\
	\Psi_i(\beta)
	&=
	-
	\frac{\partial^2}{\partial \beta^2}
	\log p(x_i \mid \eta_i, \hat{\beta}),
\end{align}
where each term quantifies the local contribution of node $i$ to the overall information geometry of the field. These local estimators form the basis for defining curvature-driven operators, such as the shape operator employed in the proposed CuBAS framework.

Substituting the conditional distribution of the Potts model into the definition of $\Phi_i(\beta)$ yields
\begin{equation}
	\Phi_i(\beta)
	=
	\left[
	U_i(x_i)
	-
	\frac{
		\sum_{\ell=1}^{q}
		U_i(\ell)\exp\{\beta U_i(\ell)\}
	}{
		\sum_{\ell=1}^{q}
		\exp\{\beta U_i(\ell)\}
	}
	\right]^2,
\end{equation}
where $U_i(\ell)$ denotes the local energy associated with assigning label $\ell$ to node $i$. The first term corresponds to the observed local energy, while the second term represents its expectation under the model’s conditional distribution.

This expression can be rewritten in a more compact and interpretable form as
\begin{equation}
	\Phi_i(\beta)
	=
	\left[
	\frac{
		\sum_{\ell=1}^{q}
		\big(U_i(x_i)-U_i(\ell)\big)
		\exp\{\beta U_i(\ell)\}
	}{
		\sum_{\ell=1}^{q}
		\exp\{\beta U_i(\ell)\}
	}
	\right]^2,
\end{equation}
which makes explicit how $\Phi_i(\beta)$ measures the sensitivity of the local energy landscape to perturbations in $\beta$.

Expanding the squared term leads to a symmetric formulation
\begin{equation}
	\Phi_i(\beta)
	=
	\frac{
		\sum_{\ell=1}^{q}\sum_{k=1}^{q}
		\bar{U}_i(\ell,k)
		\exp\{\beta (U_i(\ell)+U_i(k))\}
	}{
		\sum_{\ell=1}^{q}\sum_{k=1}^{q}
		\exp\{\beta (U_i(\ell)+U_i(k))\}
	},
\end{equation}
where
\[
\bar{U}_i(\ell,k)
=
\big(U_i(x_i)-U_i(\ell)\big)
\big(U_i(x_i)-U_i(k)\big).
\]
This formulation highlights the role of $\Phi_i(\beta)$ as a local dispersion measure of energy deviations, capturing the degree of statistical variability and curvature around node $i$. Such curvature-sensitive quantities are central to the decomposition of the graph into low- and high-information regions, as explored in the subsequent sections.

\subsection{Second-order Fisher information and curvature characterization}

We now derive the local \textit{second-order} Fisher information associated with the 
$q$-state Potts model, which captures curvature properties of the induced statistical 
manifold. Starting from the conditional log-likelihood and taking the second derivative 
with respect to the coupling parameter $\beta$, the local second-order Fisher information 
at node $i$ is given by
\begin{equation}
	\Psi_i(\beta)
	=
	-
	\frac{\partial^2}{\partial \beta^2}
	\log p(x_i \mid \eta_i, \beta).
\end{equation}
Evaluating this derivative yields
\begin{align}
	\Psi_i(\beta)
	&=
	\frac{
		\left[
		\sum_{\ell=1}^{q}
		U_i(\ell)^2
		\exp\{\beta U_i(\ell)\}
		\right]
		\left[
		\sum_{\ell=1}^{q}
		\exp\{\beta U_i(\ell)\}
		\right]
	}{
		\left[
		\sum_{\ell=1}^{q}
		\exp\{\beta U_i(\ell)\}
		\right]^2
	}
	\nonumber \\
	&\quad -
	\frac{
		\left[
		\sum_{\ell=1}^{q}
		U_i(\ell)
		\exp\{\beta U_i(\ell)\}
		\right]^2
	}{
		\left[
		\sum_{\ell=1}^{q}
		\exp\{\beta U_i(\ell)\}
		\right]^2
	}.
\end{align}

This expression admits a direct probabilistic interpretation: it corresponds to the 
conditional variance of the local energy function $U_i(\cdot)$ under the Gibbs 
distribution parameterized by $\beta$. As such, $\Psi_i(\beta)$ quantifies the local 
stability of the energy landscape and provides a curvature-sensitive measure complementary 
to the first-order Fisher information.

By straightforward algebraic manipulation, the above expression can be rewritten in a 
pairwise summation form,
\begin{align}
	\Psi_i(\beta)
	&=
	\frac{
		\sum_{\ell=1}^{q}\sum_{k=1}^{q}
		U_i(\ell)^2
		\exp\{\beta (U_i(\ell)+U_i(k))\}
	}{
		\left[
		\sum_{\ell=1}^{q}
		\exp\{\beta U_i(\ell)\}
		\right]^2
	}
	\nonumber \\
	&\quad -
	\frac{
		\sum_{\ell=1}^{q}\sum_{k=1}^{q}
		U_i(\ell)U_i(k)
		\exp\{\beta (U_i(\ell)+U_i(k))\}
	}{
		\left[
		\sum_{\ell=1}^{q}
		\exp\{\beta U_i(\ell)\}
		\right]^2
	}.
\end{align}

Grouping common exponential terms leads to the compact and symmetric representation
\begin{equation}
	\Psi_i(\beta)
	=
	\frac{
		\sum_{\ell=1}^{q}\sum_{k=1}^{q}
		\hat{U}_i(\ell,k)
		\exp\{\beta (U_i(\ell)+U_i(k))\}
	}{
		\sum_{\ell=1}^{q}\sum_{k=1}^{q}
		\exp\{\beta (U_i(\ell)+U_i(k))\}
	},
\end{equation}
where
\begin{equation}
	\hat{U}_i(\ell,k)
	=
	U_i(\ell)^2 - U_i(\ell)U_i(k).
\end{equation}
This formulation makes explicit that $\Psi_i(\beta)$ captures second-order interactions 
between alternative label assignments in the local neighborhood of node $i$, thereby 
encoding curvature information of the statistical manifold induced by the Potts model.

Together, the quantities $\Phi_i(\beta)$ and $\Psi_i(\beta)$ describe complementary aspects 
of the local information geometry: while $\Phi_i(\beta)$ reflects the sensitivity of the 
log-likelihood to infinitesimal perturbations in $\beta$ (metric structure), $\Psi_i(\beta)$ 
characterizes the local curvature and stability of the induced energy landscape. This dual 
description is central to the curvature-based decomposition and adaptive sampling strategy 
proposed in this work.

\paragraph{Tensorial formulation and computational aspects.}
To enable efficient computation and facilitate matrix-based implementations, we express 
both Fisher information terms using tensorial operations. Define the vectors
\begin{equation}
	\vec{v}_i =
	\begin{bmatrix}
		U_i(x_i)-U_i(1) \\
		U_i(x_i)-U_i(2) \\
		\vdots \\
		U_i(x_i)-U_i(q)
	\end{bmatrix},
	\qquad
	\vec{w}_i =
	\begin{bmatrix}
		\exp\{\beta U_i(1)\} \\
		\exp\{\beta U_i(2)\} \\
		\vdots \\
		\exp\{\beta U_i(q)\}
	\end{bmatrix}.
\end{equation}

We further introduce the $q \times q$ matrices
\begin{equation}
	A_i =
	\begin{bmatrix}
		U_i(1) & \cdots & U_i(1) \\
		U_i(2) & \cdots & U_i(2) \\
		\vdots & \ddots & \vdots \\
		U_i(q) & \cdots & U_i(q)
	\end{bmatrix},
	\qquad
	B_i =
	\begin{bmatrix}
		0 & \cdots & U_i(1)-U_i(q) \\
		\vdots & \ddots & \vdots \\
		U_i(q)-U_i(1) & \cdots & 0
	\end{bmatrix},
\end{equation}
and let $\Lambda_i = A_i \odot B_i$, where $\odot$ denotes the Hadamard (elementwise) product.

Using the Kronecker product $\otimes$, the local first- and second-order Fisher information 
terms can be compactly expressed as
\begin{align}
	\Phi_i(\beta)
	&=
	\frac{
		\left\lVert
		\left(\vec{v}_i \odot \vec{w}_i\right)
		\otimes
		\left(\vec{v}_i \odot \vec{w}_i\right)^{\!T}
		\right\rVert_{+}
	}{
		\left\lVert
		\vec{w}_i \otimes \vec{w}_i
		\right\rVert_{+}
	},
	\\[0.5em]
	\Psi_i(\beta)
	&=
	\frac{
		\left\lVert
		\Lambda_i \odot
		\left(\vec{w}_i \otimes \vec{w}_i\right)
		\right\rVert_{+}
	}{
		\left\lVert
		\vec{w}_i \otimes \vec{w}_i
		\right\rVert_{+}
	},
\end{align}
where $\lVert \cdot \rVert_{+}$ denotes the sum of all elements of its argument. This tensorial representation not only reduces computational complexity through vectorized operations but also provides a natural algebraic structure for subsequent curvature-based analysis and sampling decisions.

\paragraph{Shape operator}
To assess the local geometric structure induced by the Potts model at each labeled sample, 
we construct the \textit{local shape operator}, a fundamental object in differential geometry 
that encodes both intrinsic and extrinsic curvature properties of a manifold. 
The shape operator provides direct access to classical curvature measures: its determinant, 
trace, and eigenvalues correspond to the Gaussian curvature, mean curvature, and principal 
curvatures, respectively \cite{Manfredo}. In the present setting, this operator establishes 
a rigorous link between local statistical interactions in the Markov random field and the 
global geometric behavior of the associated statistical manifold.

\begin{definition}[Shape Operator]
	Let $M$ be a regular surface endowed with the first fundamental form $\mathbb{I}$ 
	and the second fundamental form $\mathbb{II}$. 
	The \textit{shape operator} $P$ is the linear map that relates variations of the unit 
	normal vector field to the local surface geometry, and is defined as \cite{Manfredo}
	\begin{equation}
		P = -\,\mathcal{II}(\theta)\,\mathcal{I}(\theta)^{-1}.
	\end{equation}
\end{definition}

At an intuitive level, it quantifies how the manifold’s orientation varies when one moves along tangent directions, namely how the unit normal vector bends locally. This viewpoint is especially attractive in computational and machine learning contexts, since the shape operator encodes second-order geometric information as a linear map acting on the tangent space. In the context of the $q$-state Potts model considered in this work, the associated statistical manifold is \textit{one-dimensional}, being parameterized solely by the inverse temperature $\beta$. As a consequence, the shape operator reduces to a scalar quantity, which directly measures the \textit{local curvature} of the statistical manifold at a given value of $\beta$. This scalar curvature captures how rapidly the local geometry of the likelihood surface bends in response to perturbations of the coupling parameter, thereby providing an intrinsic and interpretable measure of local information concentration. Motivated by this observation, we define the \textit{local scalar shape operator} at node $i$ as
\begin{equation}
	S_i(\beta)
	=
	-\,\frac{\Psi_i(\beta)}{\Phi_i(\beta) + \lambda}
	=
	-\,\frac{
		\left\lVert
		\Lambda_i \odot
		\left(
		\vec{w}_i \otimes \vec{w}_i
		\right)
		\right\rVert_{+} 
	}{
		\left\lVert
		\left(
		\vec{v}_i \odot \vec{w}_i
		\right)
		\otimes
		\left(
		\vec{v}_i \odot \vec{w}_i
		\right)^{T}
		\right\rVert_{+}
		+ \lambda
	},
\end{equation}
where $\lambda = 10^{-4}$ is a small regularization constant introduced to ensure numerical 
stability and to prevent division by zero in degenerate configurations.

The resulting quantity $S_i(\beta)$ quantifies the \textit{local curvature contribution} of 
node $i$ within the labeled graph. High absolute values of $S_i(\beta)$ correspond to regions 
of pronounced curvature, indicating neighborhoods where the local statistical structure 
changes rapidly and where the information content is highly concentrated. Conversely, 
near-zero curvature values are associated with flatter regions of the statistical manifold, 
reflecting redundancy or limited discriminative power. This curvature-driven characterization 
forms the cornerstone of the proposed dataset filtering and decomposition strategy.

\paragraph{Curvature-Based Adaptive Sampling (CuBAS)}
Building upon the above geometric formulation, we introduce the 
\textit{Curvature-Based Adaptive Sampling} (CuBAS) strategy, a principled mechanism for 
selecting informative samples from labeled datasets through the lens of information geometry. 
By interpreting the data distribution as a statistical surface induced by the Potts model, 
CuBAS exploits curvature information derived from the local shape operator to identify regions of high and low information density. In essence, the method adaptively decomposes the dataset into complementary subsets that reflect distinct geometric regimes of the underlying 
information field. The complete procedure consists of six main steps, summarized in  Algorithm~\ref{alg:cubas}.

\begin{algorithm}
	\caption{CuBAS: Curvature-Based Adaptive Data Decomposition}
	\label{alg:cubas}
	\begin{algorithmic}[1]
		\Require 
		Labeled dataset $\mathcal{D} = \{(x_i, y_i)\}_{i=1}^{n}$; 
		distance function $d(\cdot,\cdot)$; 
		number of neighbors $k$; 
		information threshold $T \in [0,1]$
		percentage of samples in training set $p$ 
		\Ensure 
		Low-information node set $\mathcal{L}$ and high-information node set $\mathcal{H}$
		
		\State \textbf{Graph construction:}
		Construct a $k$-nearest neighbors (k-NN) graph 
		$G = (\mathcal{V}, \mathcal{E})$ from $\mathcal{D}$ using distance $d(\cdot,\cdot)$.
		
		\State \textbf{Parameter estimation:}
		Estimate the inverse temperature parameter $\beta$ of the Potts model on $G$ 
		via maximum pseudo-likelihood (MPL).
		
		\State \textbf{Local curvature estimation:}
		\For{each node $i \in \mathcal{V}$}
		\State Compute the local first-order Fisher information $\Phi_i(\beta)$.
		\State Compute the local second-order Fisher information $\Psi_i(\beta)$.
		\State Compute the local scalar shape operator:
		\[
		S_i(\beta) \leftarrow -\frac{\Psi_i(\beta)}{\Phi_i(\beta) + \lambda}.
		\]
		\EndFor
		
		\State \textbf{Curvature normalization:}
		Normalize $\{S_i(\beta)\}_{i=1}^{n}$ to the interval $[0,1]$.
		
		\State \textbf{Information-based partitioning:} Define the low and high information samples
		\State Partition the graph into
		\[
		\mathcal{L}=\{i:S_i(\beta)<T\},
		\qquad
		\mathcal{H}=\{i:S_i(\beta)\ge T\}.
		\]
		\State \textbf{Sampling strategy:} Choose samples from $\mathcal{L}$ and $\mathcal{H}$
		\State Compute the number of samples required from $\mathcal{L}$ and $\mathcal{H}$:
		\[
		\mathcal{N}_{L}=\lfloor p|\mathcal{L}| \rfloor
		\]
		\[
		\mathcal{N}_{H}=\lfloor p|\mathcal{H}| \rfloor
		\]
		\State Randomly sample $\mathcal{N}_{L}$ nodes from $\mathcal{L}$ without replacement to create $\mathcal{X}_{L}$.
		\State Randomly sample $\mathcal{N}_{H}$ nodes from $\mathcal{H}$ without replacement to create $\mathcal{X}_{H}$.
		\State $\mathcal{D}_{train} \gets \left[ \mathcal{X}_L, \mathcal{X}_H \right]$	\Comment{Training set with $pn$ samples}
		\State $\mathcal{D}_{test} \gets \left[ \mathcal{L} \setminus \mathcal{X}_L, \mathcal{H} \setminus \mathcal{X}_H \right]$	\Comment{Testing set with $(1-p)~n$ samples}
		\State \textbf{return} ($\mathcal{D}_{train}$, $\mathcal{D}_{test}$)
		\end{algorithmic}
\end{algorithm}

The resulting decomposition induces a geometrically grounded criterion for 
adaptive sampling. In particular, the subset $\mathcal{H}$ isolates regions of 
high structural and informational relevance, such as class boundaries and 
heterogeneous neighborhoods, whereas $\mathcal{L}$ corresponds to smooth and 
largely redundant regions of the data manifold. By explicitly separating these 
two regimes, the proposed curvature-driven strategy achieves a favorable trade-off 
between representation compactness and information diversity, thereby improving 
the efficiency and robustness of subsequent learning stages.

The core principle underlying the proposed \textit{Curvature-Based Adaptive Sampling} 
(CuBAS) framework is the selective \textit{filtering} of labeled samples according 
to the information content encoded in their local neighborhood configurations. 
This mechanism admits a natural analogy with classical \textit{low-pass} and 
\textit{high-pass} filtering in signal and image processing. In regions exhibiting 
smooth label transitions, commonly observed in supervised classification settings 
with moderate to large inverse temperature values $\beta$, the $\mathcal{L}$-nodes 
act as a low-pass component, yielding a smoothed approximation of the original 
$k$-NN graph that preserves dominant cluster structures and intra-class coherence. 
In contrast, the $\mathcal{H}$-nodes function as a high-pass component, selectively 
highlighting high-curvature regions associated with abrupt changes in the local 
label configuration. These regions typically correspond to class interfaces and 
decision boundaries, and therefore capture the most discriminative interactions 
in the data.

By explicitly focusing on these high-curvature nodes, CuBAS identifies samples 
that are critical for defining inter-class relationships, while suppressing 
redundant samples that contribute little to the decision structure. This 
geometric filtering perspective not only provides an intuitive interpretation 
of the proposed method, but also explains its empirical effectiveness in 
enhancing classification performance under constrained sampling budgets.

\begin{figure}
	\centering
	\includegraphics[scale=0.5]{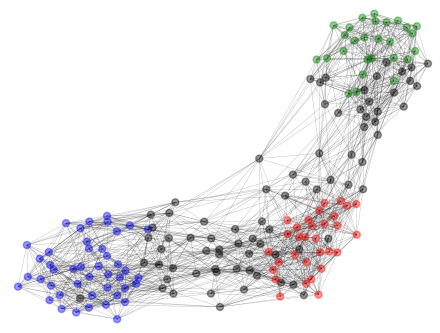}
	\includegraphics[scale=0.5]{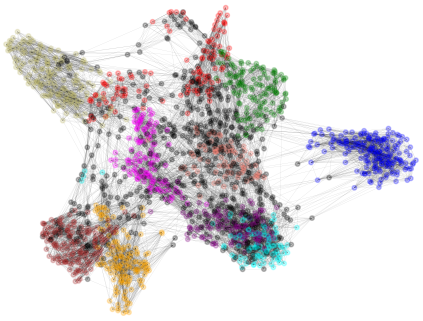}
	\includegraphics[scale=0.5]{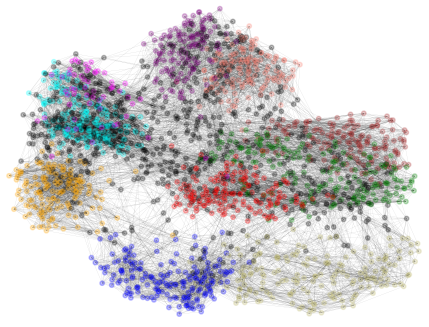}
	\includegraphics[scale=0.5]{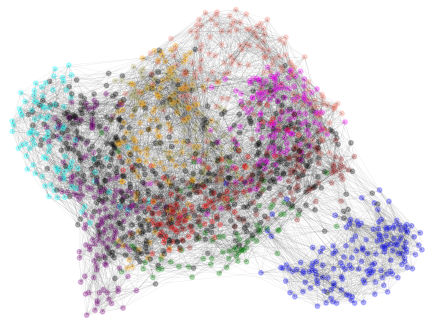}
	\caption{Visualization of high-curvature nodes ($\mathcal{H}$-nodes) identified by CuBAS in the \textit{digits}, \textit{mfeat\_karhunen}, \textit{semeion} and \textit{wine} datasets. High-curvature samples are highlighted in black and concentrate around class boundaries, where inter-class interactions are most prominent.}
	\label{fig:H}
\end{figure}

Figure~\ref{fig:H} illustrates the proposed decomposition on two representative 
datasets. The labeled $k$-NN graphs are shown with high-curvature nodes 
($\mathcal{H}$-nodes) highlighted in black. As can be observed, these nodes are 
predominantly concentrated in the vicinity of decision boundaries, confirming 
that the proposed curvature measure successfully identifies samples responsible 
for inter-class connectivity in the graph.

\subsection{Adaptive Information Threshold Estimation}
\label{subsec:adaptive_threshold}

A critical design choice in our framework is the threshold $T$ that
partitions the normalized curvature scores $K = \{K_i\}_{i=1}^{n}$,
$K_i \in [0,1]$, into low-curvature and high-curvature populations.
A fixed quantile threshold, while simple, fails to account for the
heterogeneous shape of the curvature distribution across datasets:
some labeled graphs exhibit a clearly bimodal curvature profile, with
a dense low-curvature mode corresponding to homogeneous clusters and
a separated high-curvature mode concentrated near decision boundaries,
while others exhibit a unimodal, right-skewed profile in which only a
small outlier tail is genuinely informative. We therefore propose an
\emph{adaptive} threshold estimator that selects between two
complementary criteria based on a quantitative bimodality diagnostic,
rather than committing a priori to either regime.

\subsubsection{Otsu's Criterion for Bimodal Curvature Distributions}

When the curvature distribution is bimodal, the threshold separating
the two populations can be estimated by maximizing the between-class
variance of the induced partition, following the classical criterion
of \citet{otsu1979threshold}. Given a histogram of $K$ with $B$ bins
and bin centers $\{\kappa_b\}_{b=1}^{B}$, a candidate threshold $T$
induces a low-curvature class $L(T) = \{i : K_i < T\}$ and a
high-curvature class $H(T) = \{i : K_i \geq T\}$ with population
fractions $w_L(T)$ and $w_H(T) = 1 - w_L(T)$, and class means
$\mu_L(T)$ and $\mu_H(T)$. The optimal threshold maximizes the
inter-class variance

\begin{equation}
	\label{eq:otsu_criterion}
	T_{\mathrm{Otsu}}^{*} = \operatorname*{arg\,max}_{T \in [0,1]}
	\; \sigma_B^2(T), \qquad
	\sigma_B^2(T) = w_L(T)\, w_H(T) \, \left[\mu_L(T) - \mu_H(T)\right]^2 .
\end{equation}

We compute $\sigma_B^2(T)$ efficiently over all candidate bin
boundaries using cumulative sums of the histogram counts and weighted
counts, yielding an exact $O(B)$ evaluation rather than the naive
$O(B^2)$ search over class pairs. If the curvature histogram is
degenerate (i.e., $\max_T \sigma_B^2(T) \approx 0$, indicating no
discriminative bin boundary exists), the Otsu criterion is flagged as
unreliable and the fallback procedure described in
Section~\ref{subsubsec:iqr_fallback} is invoked.

\subsubsection{Bimodality Validation via Ashman's $D$ Coefficient}

Because Otsu's criterion always returns \emph{some} threshold
regardless of whether the underlying distribution is genuinely
bimodal, we validate its applicability using Ashman's $D$ statistic
\citep{ashman1994detecting}, a standard measure of separation between
two Gaussian components. We fit a two-component Gaussian mixture
model (GMM) to $K$,

\begin{equation}
	\label{eq:gmm_fit}
	p(K_i) = \pi_1 \, \mathcal{N}(K_i \mid \mu_1, \sigma_1^2)
	+ \pi_2 \, \mathcal{N}(K_i \mid \mu_2, \sigma_2^2),
\end{equation}

via expectation-maximization with multiple random restarts, and
compute

\begin{equation}
	\label{eq:ashman_D}
	D = \frac{\sqrt{2}\,\lvert \mu_1 - \mu_2 \rvert}
	{\sqrt{\sigma_1^2 + \sigma_2^2}} .
\end{equation}

The coefficient $D$ quantifies the separation between the two
estimated modes relative to their dispersion: $D \geq 2$ is the
classical threshold in the mixture-modeling literature for
\emph{well-separated, unimodally-resolvable} components, below which
the two-component fit cannot be reliably distinguished from a single
mode \citep{ashman1994detecting}. In our implementation we adopt a
slightly more conservative threshold, $D_{\min} = 2.6$, to reduce the
risk of spuriously invoking the bimodal regime on curvature
distributions with heavy but unimodal tails. Additionally, since a
two-component GMM can in principle converge to two modes that are
both concentrated near zero curvature (i.e., $\mu_2 < 0.1$, indicating
the ``high-curvature'' component is itself an artifact of noise in a
predominantly homogeneous graph), we treat this configuration as a
degenerate case and revert to the fallback criterion even when
$D \geq D_{\min}$.

We note that Ashman's $D$ statistic, as originally formulated,
implicitly assumes comparable dispersion between the two mixture
components. When the two curvature modes are markedly heteroscedastic
($\sigma_1 \neq \sigma_2$), the asymptotic $\chi^2$ behavior underlying
related likelihood-ratio bimodality tests breaks down for small or
moderate sample sizes, and a parametric bootstrap is recommended
instead \citep{lo2008likelihood}. In our setting, the curvature
populations induced by $\mathcal{L}$ and $\mathcal{H}$ are not assumed
homoscedastic; we therefore treat $D \geq D_{\min}$ as a practical
screening criterion rather than a formal hypothesis test, consistent
with its standard usage in the mixture-modeling literature
\citep{ashman1994detecting,lo2008likelihood}.

\subsubsection{Tukey-Fence Fallback for Unimodal Distributions}
\label{subsubsec:iqr_fallback}

When the bimodality test fails ($D < D_{\min}$) or the Otsu criterion
itself is degenerate, the curvature distribution is treated as
effectively unimodal, and only the upper outlier tail is considered
genuinely high-information. We adopt a quantile-based threshold
informed by the interquartile range (IQR),

\begin{equation}
	\label{eq:iqr_fallback}
	T_{\mathrm{IQR}} =
	\begin{cases}
		Q_3(K), & \text{if } \mathrm{IQR}(K) > 0, \\[4pt]
		Q_{0.95}(K), & \text{if } \mathrm{IQR}(K) = 0,
	\end{cases}
\end{equation}

where $Q_3(K)$ is the third quartile and $\mathrm{IQR}(K) = Q_3(K) -
Q_1(K)$. The degenerate case $\mathrm{IQR}(K) = 0$ arises when the
curvature distribution is heavily concentrated (e.g., more than $75\%$
of nodes share an identical curvature value, typically $K_i = 0$ after
normalization), in which case the $75^{\mathrm{th}}$-percentile
threshold collapses and the $95^{\mathrm{th}}$-percentile is used
instead to preserve a non-trivial high-curvature subset.

\subsubsection{Decision Cascade}

The complete adaptive threshold estimator is summarized as a decision
cascade in Algorithm~\ref{alg:adaptive_threshold}. The procedure first
attempts the bimodal (Otsu) criterion and accepts it only if both (i)
the inter-class variance is non-degenerate and (ii) Ashman's $D$
statistic confirms genuine separation between the two curvature modes
with the high-curvature mode being non-trivial ($\mu_2 \geq 0.1$).
Otherwise, the method falls back to the Tukey-fence quantile rule of
Eq.~\eqref{eq:iqr_fallback}. This cascade allows $T$ to adapt not only
to the dataset at hand, but implicitly to the local geometry and
class overlap structure encoded in the Potts-derived curvature field,
without requiring any dataset-specific tuning.

\begin{algorithm}[t]
	\caption{Adaptive Threshold Estimation}
	\label{alg:adaptive_threshold}
	\begin{algorithmic}[1]
		\Require Normalized curvature scores $K \in [0,1]^n$, number of histogram bins $B$, bimodality threshold $D_{\min}$
		\Ensure Threshold $T^{*}$, selection method $\mathrm{tag} \in \{\textsc{Otsu}, \textsc{IQR}, \textsc{Degenerate}\}$
		\If{$\max(K) - \min(K) < \epsilon$}
		\State \Return $Q_{0.95}(K)$, $\textsc{Degenerate}$
		\EndIf
		\State Compute $T_{\mathrm{Otsu}}$ via Eq.~\eqref{eq:otsu_criterion} using cumulative-sum histogram evaluation
		\State Fit two-component GMM to $K$; compute $D$, $\mu_1$, $\mu_2$ via Eqs.~\eqref{eq:gmm_fit}--\eqref{eq:ashman_D}
		\If{$T_{\mathrm{Otsu}}$ is defined \textbf{and} $D \geq D_{\min}$}
		\If{$\min(\mu_1, \mu_2)$'s high-curvature component $\mu_2 < 0.1$}
		\State $T^{*} \gets T_{\mathrm{IQR}}$ via Eq.~\eqref{eq:iqr_fallback} \Comment{spurious bimodal fit; revert to fallback}
		\State \Return $T^{*}$, $\textsc{Degenerate}$
		\Else
		\State \Return $T_{\mathrm{Otsu}}$, $\textsc{Otsu}$
		\EndIf
		\Else
		\State $T^{*} \gets T_{\mathrm{IQR}}$ via Eq.~\eqref{eq:iqr_fallback}
		\State \Return $T^{*}$, $\textsc{IQR}$
		\EndIf
	\end{algorithmic}
\end{algorithm}

Once $T^{*}$ is determined, the high- and low-information partitions
used throughout Section~\ref{sec:method} are obtained directly as

\begin{equation}
	\label{eq:HL_partition_adaptive}
	\mathcal{H} = \{ i : K_i \geq T^{*} \}, \qquad
	\mathcal{L} = \{ i : K_i < T^{*} \},
\end{equation}

replacing the fixed-quantile rule used in our preliminary experiments
with a threshold that responds to the empirical shape of the
curvature field on a per-dataset basis.

\subsection{Computational Complexity}
\label{subsec:complexity}

We analyze the computational complexity of CuBAS by decomposing the
algorithm into its constituent stages: $k$-NN graph construction,
label-count matrix computation, maximum pseudo-likelihood (MPL)
estimation of $\beta$, Fisher information computation, adaptive
threshold estimation, and graph partitioning.

\paragraph{$k$-NN graph construction.}
Given a dataset of $n$ samples in $\mathbb{R}^m$, constructing the
$k$-nearest-neighbor graph requires computing pairwise distances and
retrieving the $k$ closest neighbors for each point. Using a
$k$-d tree or ball tree index, this step runs in $O(n m \log n)$
for low-to-moderate dimensions and degrades to $O(n^2 m)$ in the
high-dimensional regime where tree-based acceleration is ineffective. The resulting graph has $nk$ edges and is stored as a sparse $n \times n$ matrix in compressed
sparse row (CSR) format, requiring $O(nk)$ memory.

\paragraph{Label-count matrix.}
The label-count matrix $\mathbf{U} \in \mathbb{Z}^{n \times c}$,
where $U_{ij}$ counts the number of neighbors of node $i$ carrying
label $j$, is computed in a single pass over the $nk$ edges of the
graph. This step is $O(nk)$ in time and $O(nc)$ in memory, and
constitutes a one-time precomputation shared by all subsequent stages.

\paragraph{Maximum pseudo-likelihood estimation.}
The MPL estimator $\beta_{\mathrm{MPL}}$ is obtained as the root of
the scalar equation $\partial \ell_{\mathrm{PL}} / \partial \beta = 0$,
solved via the secant method. Each evaluation of the pseudo-likelihood
gradient requires computing, for each of the $n$ nodes, a softmax-like
normalization over $c$ classes, yielding a per-iteration cost of
$O(nc)$. The secant method converges superlinearly and requires a small
number of iterations in practice (typically fewer than 20), so the
total cost of MPL estimation is $O(T_{\beta} \, nc)$, where $T_{\beta}$
denotes the number of root-finder iterations.

\paragraph{Fisher information.}
Given $\beta_{\mathrm{MPL}}$ and the precomputed matrix $\mathbf{U}$,
the first- and second-order local Fisher information scores $\Phi_i$
and $\Psi_i$ are computed for all $n$ nodes simultaneously via
vectorized NumPy operations over the $(n \times c)$ arrays
$\mathbf{U}$, $\mathbf{W} = \exp(\beta \mathbf{U})$, and their
elementwise products. Each of the required reductions (row-wise sums,
squared sums, and weighted sums) is $O(nc)$, so the total cost of
this stage is $O(nc)$. The curvature scores $K_i = -\Psi_i /
(\Phi_i + \varepsilon)$ are then obtained in $O(n)$.

\paragraph{Adaptive threshold estimation.}
The Otsu criterion is evaluated over a histogram of $K$ with $B$ bins
using cumulative sums, at a cost of $O(n + B)$. Fitting the
two-component GMM for the Ashman $D$ diagnostic requires
$O(T_{\mathrm{EM}} \, nc_{\mathrm{GMM}})$, where $T_{\mathrm{EM}}$
is the number of EM iterations and $c_{\mathrm{GMM}} = 2$ is fixed.
Both costs are negligible relative to the preceding stages for
typical values of $n$ and $B$.

\paragraph{Graph partitioning and subset selection.}
Thresholding the curvature scores to obtain $\mathcal{H}$ and
$\mathcal{L}$ is $O(n)$. Sampling $\lfloor \rho |\mathcal{H}| \rfloor$
nodes from $\mathcal{H}$ and $\lfloor \rho |\mathcal{L}| \rfloor$
nodes from $\mathcal{L}$ at a given training fraction $\rho \in (0,1)$
is $O(n)$ via reservoir sampling.

\paragraph{Overall complexity.}
Summing over all stages, the dominant costs are $k$-NN graph
construction and Fisher information computation. The total time
complexity of CuBAS is

\begin{equation}
	\label{eq:complexity}
	O\!\left(nm\log n + nk + T_{\beta}\,nc\right) ,
\end{equation}

which reduces to $O(nm \log n)$ when $k$, $c$, and $T_{\beta}$ are
treated as constants relative to $n$ and $m$, the typical regime
in supervised classification benchmarks. Crucially, once the $k$-NN
graph is constructed and $\beta_{\mathrm{MPL}}$ is estimated, the
Fisher information, curvature scores, and graph partition are computed
in $O(nc)$, i.e., \emph{linear in the number of samples and classes}.
This means that the curvature decomposition itself, the
distinguishing computational contribution of CuBAS relative to
graph construction, which is shared with any $k$-NN-based method,
adds only a linear overhead to the pipeline. Table~\ref{tab:complexity}
summarizes the per-stage complexity.

\begin{table}
	\centering
	\caption{Per-stage time complexity of CuBAS. $n$: number of samples;
		$m$: number of features; $k$: number of nearest neighbors; $c$:
		number of classes; $T_{\beta}$: root-finder iterations (typically
		$< 20$); $B$: histogram bins for Otsu (default $256$).}
	\label{tab:complexity}
	\begin{tabular}{lll}
		\toprule
		\textbf{Stage} & \textbf{Time complexity} & \textbf{Dominant regime} \\
		\midrule
		$k$-NN graph construction  & $O(nm \log n)$         & Low/moderate $m$         \\
		& $O(n^2 m)$             & High $m$ (HDLSS)         \\
		Label-count matrix $\mathbf{U}$ & $O(nk)$           & Always                   \\
		MPL estimation ($\beta$)   & $O(T_{\beta}\,nc)$     & Always                   \\
		Fisher information         & $O(nc)$                & Always                   \\
		Curvature scores           & $O(n)$                 & Always                   \\
		Adaptive threshold         & $O(n + B)$             & Always                   \\
		Graph partitioning         & $O(n)$                 & Always                   \\
		\midrule
		\textbf{Total}             & $O(nm\log n + T_{\beta}\,nc)$ & Low/moderate $m$  \\
		\bottomrule
	\end{tabular}
\end{table}

\paragraph{Comparison with baseline methods.}
Random sampling incurs no preprocessing cost beyond $O(n)$ for index
generation. Entropy-based uncertainty sampling requires fitting an
auxiliary classifier on the seed set and evaluating its posterior over
the unlabeled pool; for a $k$-NN classifier this costs $O(n_s m k_s +
n_p k_s)$, where $n_s$ and $n_p$ are the seed and pool sizes and
$k_s$ the number of neighbors of the auxiliary classifier. The one-time,
model-free preprocessing profile makes CuBAS particularly attractive
in settings where repeated querying or auxiliary model training would
be prohibitively expensive.

\section{Computational Experiments and Results}
\label{sec:experiments}
This section presents a comprehensive empirical evaluation of CuBAS, designed to address three research questions. We begin by describing the experimental protocol, then compare
CuBAS against random sampling and a representative classical uncertainty-based baseline (entropy-based sampling) across multiple real datasets. Finally, discuss the conditions under which curvature-aware sampling yields the largest gains over geometry-agnostic
alternatives. Taken together, these experiments provide consistent evidence that exploiting the local geometry of the labeled data manifold, as captured by the observed Fisher information under the Potts model, leads to more informative and more compact training subsets than purely uncertainty-driven or random selection strategies. In this work, we seek to answer three fundamental questions:
\vspace{-0.25cm}
\begin{enumerate}
	\item Does curvature-aware sampling via CuBAS select training subsets that are more informative than those obtained by classical uncertainty-based and geometry-agnostic sampling strategies?
	\item How does the benefit of curvature-aware sampling scale with training set size, number of features and classes?
	\item Does the adaptive threshold estimation strategy for $T$ handle the robustness and performance of CuBAS across datasets with heterogeneous curvature distributions?
\end{enumerate}
\vspace{-0.1cm}
Experiments were conducted on a diverse collection of more than 60 publicly available datasets obtained from OpenML \citep{vanschoren2013openml,bischl2021openml}. The benchmark spans a broad range of sample sizes, numbers of attributes, and numbers of classes, encompassing both balanced and imbalanced classification problems as well as datasets with varying degrees of class overlap and cluster structure. This diversity enables the evaluation of CuBAS across the wide range of conditions encountered in modern supervised classification tasks. All datasets, preprocessing scripts, and experimental results are publicly available to support reproducibility. Tables \ref{tab:small}, \ref{tab:large} and \ref{tab:high} show the datasets divided in three categories: small ($n < 1100$), large ($n > 2000$) and high-dimensional ($m > 1000$).


The benchmark suite was deliberately assembled to encompass a broad range of
classification scenarios, including problems with different sample sizes,
feature dimensionalities, numbers of classes, and application domains.
Such diversity is essential for assessing whether the proposed curvature-based
sampling strategy generalizes across distinct data distributions rather than
being tailored to a particular family of datasets.

Tables~\ref{tab:small}--\ref{tab:high} summarize the datasets employed in the
experimental evaluation.
For convenience, they are grouped into three categories according to their
scale and dimensionality.
Table~\ref{tab:small} contains $20$ small and medium-sized datasets,
comprising up to approximately one thousand samples.
These datasets cover classical benchmark problems frequently used in the
machine learning literature, including medical diagnosis, bioinformatics,
pattern recognition, industrial monitoring, and game-related classification.
Although relatively modest in size, they exhibit substantial variability in
their intrinsic complexity, with the number of attributes ranging from only
$4$ to $856$ and the number of classes varying from binary to multiclass
problems with up to $13$ categories.

The second group, reported in Table~\ref{tab:large}, consists of $32$
large-scale datasets containing between approximately $1.2 \times 10^3$
and $1.5 \times 10^4$ instances.
These datasets include several well-established image recognition,
remote sensing, handwriting recognition, speech processing,
human activity recognition, hyperspectral imaging, and tabular classification
benchmarks.
Their feature dimensionality ranges from as few as $5$ variables
(\textit{phoneme}) to $784$ features
(\textit{MNIST}, \textit{Fashion-MNIST}, and
\textit{Kuzushiji-MNIST}), while the number of classes varies from
binary classification to problems involving $26$ distinct categories
(\textit{isolet}).
This collection enables the evaluation of CuBAS under realistic large-scale
settings where both computational efficiency and sample selection become
particularly important.

Finally, Table~\ref{tab:high} presents $16$ high-dimensional datasets,
many of which correspond to gene-expression analysis, face recognition,
mass spectrometry, and text categorization.
These datasets represent particularly challenging learning scenarios due to the
well-known ``large-$p$, small-$n$'' regime, where the number of features
substantially exceeds the number of available observations.
The dimensionality reaches more than $22\,000$ attributes
(\textit{GLI}), while several datasets contain fewer than one hundred
samples.
Such settings are especially relevant for evaluating CuBAS because redundant
or uninformative observations may significantly affect classifier
generalization, making the identification of informative samples considerably
more valuable than in conventional low-dimensional problems.

Overall, the benchmark suite spans sample sizes from only $60$ observations
(\textit{CNS}) to nearly $15\,000$ instances
(\textit{eeg-eye-state}), feature dimensionalities ranging from
$4$ to $22\,283$, and classification tasks involving between
$2$ and $40$ classes.
This considerable heterogeneity provides a rigorous experimental protocol for
assessing both the robustness and the generalization capability of the proposed
curvature-driven sampling strategy across problems exhibiting markedly
different statistical and geometric characteristics.

\begin{table}
	\centering
	\caption{Summary of the small datasets (less than 1100 samples) used in the computational experiments. Reported are the number of samples ($n$), number of features ($m$), and number of classes ($c$).}
	\begin{tabular}{ccccc}
		\toprule
		\textbf{\#} & \textbf{Dataset} & \textbf{$n$} & \textbf{$m$} & \textbf{$c$} \\
		\midrule
		1  & iris                   & 150  & 4   & 3  \\
		2  & wine                   & 178  & 13  & 3  \\
		3  & breast\_cancer         & 569  & 30  & 2  \\
		4  & Engine1         		& 383  & 5   & 3  \\
		5  & monks-problem-1        & 556  & 6   & 2  \\
		6  & credit-g         		& 1000 & 20  & 2  \\
		7  & seeds         			& 210  & 7   & 3  \\
		8  & user-knowledge        	& 403  & 5   & 5  \\
		9  & tic-tac-toe         	& 958  & 9   & 2  \\
		10  & parkinsons         	& 195  & 22  & 2  \\
		11  & diabetes         		& 768  & 8   & 2  \\
		12  & anneal         		& 898  & 38  & 5  \\
		13  & vehicle         		& 846  & 18  & 4  \\
		14  & sonar         		& 208  & 60  & 2  \\
		15  & blood-transfusion-service-center & 748 & 4 & 2 \\
		16  & tecator         		& 240  & 124 & 2  \\		
		17  & qsar-biodeg      		& 1055 & 41  & 2  \\		
		18  & arrhythmia       		& 452  & 279 & 13 \\		
		19  & parkinson-speech-uci 	& 756  & 753 & 2  \\		
		20  & cnae-9         		& 1080  &  856  & 9 \\		
		\bottomrule  
	\end{tabular}
	\label{tab:small}
\end{table}

\begin{table}
	\centering
	\caption{Summary of the large datasets (more than 2000 samples) used in the computational experiments. Reported are the number of samples ($n$), number of features ($m$), and number of classes ($c$).}
	\begin{tabular}{ccccc}
		\toprule
		\textbf{\#} & \textbf{Dataset} & \textbf{$n$} & \textbf{$m$} & \textbf{$c$} \\
		\midrule
		1  & digits									& 1797 & 64 & 10 \\
		2  & semeion								& 1593 & 256 & 10 \\
		3  & texture								& 5500 & 40 & 11 \\
		4  & segment								& 2310 & 19 & 7 \\
		5  & mfeat-morphological					& 2000 & 6 & 10 \\
		6  & mfeat-fourier							& 2000 & 76 & 10 \\
		7  & mfeat-factors							& 2000 & 216 & 10 \\
		8  & mfeat-pixel							& 2000 & 240 & 10 \\
		9  & satimage								& 6430 & 36  & 6 \\
		10  & MNIST\_784 (20\%)						& 14000 & 784 & 10 \\
		11  & Fashion-MNIST (20\%)					& 14000 & 784 & 10 \\
		12  & Kuzushiji-MNIST (20\%)				& 14000 & 784 & 10  \\
		13  & madelon								& 2600 & 500 & 2 \\
		14  & page-blocks							& 5473 & 10 & 5 \\
		15  & Indian\_pines							& 9144 & 220 & 8 \\
		16  & phoneme								& 5404 & 5 & 2 \\
		17  & sylvine								& 5124 & 20 & 2 \\
		18  & eye\_movements						& 10936 & 27 & 3 \\
		19  & GesturePhaseSegmentationProcessed		& 9873 & 32 & 5 \\
		20  & splice								& 3190 & 60 & 3 \\
		21  & hill-valley							& 1212 & 100 & 2 \\
		22  & thyroid-dis							& 2800 & 26 & 5 \\
		23  & wine-quality-white					& 4898 & 11 & 7 \\
		24  & wine-quality-red						& 1599 & 11 & 6 \\
		25  & eeg-eye-state							& 14980 & 14 & 2 \\
		26  & spambase								& 4601 & 57 & 2 \\
		27  & scene									& 2407 & 299 & 2 \\
		28  & har									& 10299 & 561 & 6 \\
		29  & nursery								& 12960 & 8 & 5 \\
		30  & isolet								& 7797 & 617 & 26 \\
		31  & ipums\_la\_99-small					& 8844 & 60 & 7 \\
		32  & waveform-5000							& 5000 & 40 & 3 \\						
		\bottomrule  
	\end{tabular}
	\label{tab:large}
\end{table}

\begin{table}
	\centering
	\caption{Summary of the high-dimensional datasets (more than 1000 features) used in the computational experiments. Reported are the number of samples ($n$), number of features ($m$), and number of classes ($c$).}
	\begin{tabular}{ccccc}
		\toprule
		\textbf{\#} & \textbf{Dataset} & \textbf{$n$} & \textbf{$m$} & \textbf{$c$} \\
		\midrule
		1  & micro-mass									& 360 & 1300 & 10 \\
		2  & UMIST\_Faces\_Cropped						& 575 & 10304 & 20 \\
		3  & Olivetti\_Faces							& 400 & 4096 & 40 \\
		4  & AP\_Lung\_Kidney							& 386 & 10935 & 2 \\
		5  & AP\_Omentum\_Kidney						& 337 & 10935 & 2 \\
		6  & AP\_Breast\_Uterus							& 468 & 10935 & 2 \\
		7  & AP\_Ovary\_Lung							& 324 & 10935 & 2 \\
		8  & AP\_Ovary\_Kidney							& 458 & 10935 & 2 \\
		9  & leukemia									& 72 & 7129 & 2 \\
		10  & GLI										& 85 & 22283 & 2 \\
		11  & MLL										& 72 & 12582 & 3 \\
		12  & CNS										& 60 & 7129 & 2 \\
		13  & arcene									& 200 & 10000 & 2 \\
		14  & tr11.wc									& 414 & 6429 & 9 \\
		15  & DLBCL										& 77 & 5469 & 2 \\
		16  & OVA\_Breast								& 1545 & 10935 & 2 \\
		\bottomrule  
	\end{tabular}
	\label{tab:high}
\end{table}

As a strong baseline for comparison, we also considered \emph{entropy-based uncertainty sampling}, one of the most widely adopted strategies in active learning \cite{Roy2001,Settles2009,ren2021survey}. Unlike random sampling, uncertainty sampling explicitly seeks the most informative training instances by selecting those for which a reference classifier exhibits the highest predictive uncertainty. Among the various uncertainty measures proposed in the literature, predictive entropy is arguably the most principled and broadly applicable, as it quantifies the dispersion of the posterior class distribution and reaches its maximum when the classifier is maximally uncertain about the correct label \cite{Shannon1948,Settles2009,ren2021survey}. This criterion has demonstrated competitive performance across a wide range of classification tasks and has become a standard benchmark in active learning research due to its simplicity, effectiveness, and model-agnostic formulation \cite{Settles2009,ren2021survey}. Comparing CuBAS against entropy-based sampling is therefore particularly informative, since both methods aim to prioritize informative samples, but rely on fundamentally different principles. Whereas entropy sampling estimates informativeness from the predictions of a trained classifier, making it inherently model-dependent, CuBAS derives informativeness directly from the geometric structure of the labeled data through curvature analysis on the induced graph. This comparison allows us to assess whether a geometry-driven, classifier-independent sampling strategy can match or surpass one of the most established uncertainty-based approaches in the active learning literature. Hence, CuBAS is not competing with weak random sampling alone, but with one of the strongest and most widely accepted information-driven sampling strategy.

The experimental protocol was designed to provide a statistically robust evaluation of the proposed sampling strategy under different levels of training data availability. For each dataset, we considered $15$ training set proportions ranging from $10\%$ to $80\%$ of the available samples, using increments of $5\%$. For every training set size, three sampling strategies were evaluated: conventional random sampling, entropy-based uncertainty sampling, and the proposed \textit{Curvature-Based Adaptive Sampling} (CuBAS) with $k = 15$ for the k-NN graph.

To account for the variability induced by the sampling process, each strategy was independently repeated $100$ times, yielding $100$ distinct training--testing partitions for every dataset and training set size. A $k$-nearest neighbor classifier ($k=5$) was then trained and evaluated on each partition using balanced accuracy as the performance metric. For each sampling strategy, the balanced accuracies obtained over the $100$ repetitions were averaged to produce a reliable performance estimate for the corresponding training set size.

Finally, to provide a global comparison between sampling strategies, we computed the mean and standard deviation of the balanced accuracy across the $15$ training set sizes. Consequently, each reported result summarizes the classifier performance over a total of $1,500$ independent training and evaluation runs per dataset and sampling strategy. The corresponding experimental results are presented in Tables~\ref{tab:results1}--\ref{tab:results3}. 

\begin{table}
	\centering
	\caption{Average balanced accuracies achieved by the k-NN classifier under random sampling, entropy-based uncertainty sampling, and the proposed CuBAS method. The evaluation was performed over 15 training set sizes (10\%--80\%, with 5\% increments). For each training size, 100 independent experiments were conducted using different train-test partitions. The reported results represent the average over the 100 repetitions and all training set sizes.}
	\begin{tabular}{cccc}
		\toprule
		\textbf{Datasets}    & \textbf{Random} & \textbf{Entropy} & \textbf{CuBAS}           \\
		\midrule
		iris                 & 0.9256 ± 0.0414 & 0.9387 ± 0.0874  & \textbf{0.9858 ± 0.0326} \\
		wine                 & 0.9501 ± 0.0310 & 0.9752 ± 0.0281  & \textbf{0.9929 ± 0.0199} \\
		breast\_cancer       & 0.9469 ± 0.0110 & 0.9752 ± 0.0148  & \textbf{0.9996 ± 0.0007} \\
		Engine1              & 0.7149 ± 0.0850 & 0.9170 ± 0.0570  & \textbf{0.9868 ± 0.0178} \\
		monks-problem-1      & 0.7743 ± 0.0826 & 0.7682 ± 0.0979  & \textbf{0.8713 ± 0.0744} \\
		credit-g             & 0.5960 ± 0.0160 & 0.5854 ± 0.0118  & \textbf{0.7940 ± 0.0616} \\
		seeds                & 0.9139 ± 0.0131 & 0.9527 ± 0.0376  & \textbf{0.9823 ± 0.0045} \\
		user-knowledge       & 0.5382 ± 0.0538 & 0.5337 ± 0.0820  & \textbf{0.6455 ± 0.0932} \\
		tic-tac-toe          & 0.7097 ± 0.0642 & 0.6839 ± 0.0831  & \textbf{0.9062 ± 0.0861} \\
		parkinsons           & 0.7713 ± 0.0784 & 0.8155 ± 0.1273  & \textbf{0.9575 ± 0.0699} \\
		diabetes             & 0.6740 ± 0.0117 & 0.6882 ± 0.0162  & \textbf{0.8599 ± 0.0360} \\
		anneal               & 0.6676 ± 0.1066 & 0.6800 ± 0.1021  & \textbf{0.8261 ± 0.0991} \\
		sonar                & 0.6735 ± 0.0385 & 0.7066 ± 0.0657  & \textbf{0.7568 ± 0.0488} \\
		vehicle              & 0.7326 ± 0.0538 & 0.7118 ± 0.1292  & \textbf{0.8825 ± 0.0730} \\
		blood-transfusion    & 0.6088 ± 0.0075 & 0.5777 ± 0.0262  & \textbf{0.6805 ± 0.0144} \\
		tecator              & 0.7756 ± 0.0592 & 0.8042 ± 0.0883  & \textbf{0.9762 ± 0.0521} \\
		qsar-biodeg          & 0.8263 ± 0.0165 & 0.8640 ± 0.0426  & \textbf{0.9948 ± 0.0101} \\
		arrhythmia           & 0.1439 ± 0.0356 & 0.1227 ± 0.0109  & \textbf{0.1724 ± 0.0434} \\
		parkinson-speech-uci & 0.7099 ± 0.0429 & 0.7142 ± 0.0358  & \textbf{0.9991 ± 0.0018} \\
		cnae-9               & 0.7499 ± 0.1054 & 0.8162 ± 0.1459  & \textbf{0.8406 ± 0.1167} \\
		\midrule
		Average              & 0.7202          & 0.7416           & \textbf{0.8556}          \\
		Median               & 0.7238          & 0.7412           & \textbf{0.8944}  \\
		\bottomrule
	\end{tabular}	
	\label{tab:results1}
\end{table}

The results admit a clear and consistent reading: CuBAS achieves the highest balanced
accuracy on every single dataset in the benchmark, with an average of $0.8556$
against $0.7416$ for entropy sampling and $0.7202$ for random sampling, corresponding to absolute improvements of $+11.4$ and $+13.5$ percentage points, respectively. The median balanced accuracy tells an even stronger story: $0.8944$ for CuBAS versus $0.7412$ for entropy sampling and $0.7238$ for random sampling, confirming that the superiority of the proposed method is not driven by a small number of outlier datasets but reflects a systematic shift in the distribution of performance across the benchmark.

The gains are particularly pronounced on datasets where the underlying
class structure exhibits well-defined geometric boundaries. On
\texttt{breast\_cancer}, CuBAS achieves a near-perfect average balanced
accuracy of $0.9996 \pm 0.0007$, compared to $0.9752$ for entropy sampling and
$0.9469$ for random sampling, a result consistent with the expectation that
the Potts curvature field identifies the narrow, geometrically coherent
decision boundary between malignant and benign tissue descriptors with high
fidelity. Similarly large gains are observed on \texttt{parkinson-speech-uci}
($0.9991$ vs.\ $0.7142$ and $0.7099$), \texttt{qsar-biodeg}
($0.9948$ vs.\ $0.8640$ and $0.8263$), \texttt{Engine1}
($0.9868$ vs.\ $0.9170$ and $0.7149$), and \texttt{tecator}
($0.9762$ vs.\ $0.8042$ and $0.7756$). In these cases, the
entropy-based baseline already improves over random sampling,
suggesting that informative samples do concentrate near the decision boundary;
however, CuBAS further exploits the global geometric structure of the
labeled graph, simultaneously capturing low-curvature prototype regions and
high-curvature boundary regions, yielding substantially more compact and
representative training subsets.

A second noteworthy pattern emerges on datasets characterized by high class
overlap or intrinsic difficulty. On \texttt{credit-g}, \texttt{diabetes},
\texttt{tic-tac-toe}, \texttt{anneal}, and \texttt{blood-transfusion}, both
random sampling and entropy sampling yield relatively low balanced accuracies
(below $0.70$), yet CuBAS recovers meaningful structure in all cases, with
gains ranging from $+7.2$ percentage points (\texttt{blood-transfusion}) to
$+19.8$ percentage points (\texttt{credit-g}). This behavior is consistent
with the theoretical motivation of our framework: in regions of high class
overlap, the curvature of the statistical manifold induced by the Potts model
is elevated precisely at the samples that straddle class boundaries, and their
selective inclusion in the training set provides the classifier with the most
discriminative geometric signal available.

The \texttt{arrhythmia} dataset stands as the single notable exception to the
general trend: all three methods yield low balanced accuracies (below $0.18$),
with CuBAS achieving $0.1724$ against $0.1439$ for random sampling and $0.1227$
for entropy sampling. This dataset is well known in the benchmark literature
for its extreme class imbalance, 13 classes with heavily skewed frequencies,
and its high ambient dimensionality relative to the number of samples,
which renders the $k$-NN graph geometry unreliable and the curvature signal
noisy. The modest but consistent improvement of CuBAS over the two baselines
even in this adversarial setting suggests that the geometric decomposition
retains some discriminative value, though the overall performance ceiling for
all methods is severely limited by the dataset's intrinsic difficulty.

Regarding the comparison between entropy sampling and random sampling, the
former achieves a marginally higher average balanced accuracy ($0.7416$
vs.\ $0.7202$), but the difference is modest and inconsistent across
datasets: entropy sampling underperforms random sampling on
\texttt{monks-problem-1}, \texttt{credit-g}, \texttt{tic-tac-toe},
\texttt{blood-transfusion}, and \texttt{vehicle}, suggesting that uncertainty
scores derived from a $k$-NN posterior trained on a small seed set can be
unreliable when the initial labeled pool is insufficient to accurately
characterize the class boundary. CuBAS, by contrast, operates directly on
the graph topology and the Potts sufficient statistics without requiring an
auxiliary classifier for scoring, and therefore does not inherit this
sensitivity to seed set quality, a structural advantage that is reflected
in its consistently lower standard deviations across most datasets.

\begin{table}
	\centering
	\caption{Average balanced accuracies achieved by the k-NN classifier under random sampling, entropy-based uncertainty sampling, and the proposed CuBAS method. The evaluation was performed over 15 training set sizes (10\%--80\%, with 5\% increments). For each training size, 100 independent experiments were conducted using different train-test partitions. The reported results represent the average over the 100 repetitions and all training set sizes.}
	\begin{tabular}{cccc}
		\toprule
		\textbf{Datasets}        & \textbf{Random} & \textbf{Entropy} & \textbf{CuBAS}           \\
		\midrule
		digits                   & 0.9534 ± 0.0249 & 0.9776 ± 0.0269  & \textbf{0.9913 ± 0.0114} \\
		semeion                  & 0.8677 ± 0.0430 & 0.9191 ± 0.0612  & \textbf{0.9488 ± 0.0364} \\
		texture                  & 0.9684 ± 0.0148 & 0.9870 ± 0.0147  & \textbf{0.9970 ± 0.0056} \\
		segment                  & 0.9167 ± 0.0247 & 0.9589 ± 0.0433  & \textbf{0.9854 ± 0.0116} \\
		mfeat-morphological      & 0.7077 ± 0.0090 & 0.7525 ± 0.0340  & \textbf{0.7985 ± 0.0101} \\
		mfeat-fourier            & 0.7651 ± 0.0351 & 0.8108 ± 0.0584  & \textbf{0.8526 ± 0.0402} \\
		mfeat-factors            & 0.9515 ± 0.0160 & 0.9812 ± 0.0215  & \textbf{0.9946 ± 0.0046} \\
		mfeat-pixel              & 0.9624 ± 0.0159 & 0.9876 ± 0.0193  & \textbf{0.9978 ± 0.0046} \\
		satimage                 & 0.8694 ± 0.0162 & 0.9280 ± 0.0335  & \textbf{0.9764 ± 0.0056} \\
		MNIST\_784 (20\%)        & 0.8935 ± 0.0201 & 0.9483 ± 0.0353  & \textbf{0.9842 ± 0.0086} \\
		Fashion-MNIST (20\%)     & 0.8088 ± 0.0174 & 0.8790 ± 0.0524  & \textbf{0.9252 ± 0.0152} \\
		Kuzushiji-MNIST (20\%)   & 0.8730 ± 0.0320 & 0.9474 ± 0.0491  & \textbf{0.9769 ± 0.0152} \\
		madelon                  & 0.5406 ± 0.0038 & 0.5475 ± 0.0097  & \textbf{0.6357 ± 0.0247} \\
		page-blocks              & 0.6975 ± 0.0767 & 0.7256 ± 0.0923  & \textbf{0.8716 ± 0.0950} \\
		Indian\_pines            & 0.7015 ± 0.0266 & 0.7030 ± 0.0377  & \textbf{0.8462 ± 0.0555} \\
		phoneme                  & 0.8105 ± 0.0239 & 0.8608 ± 0.0463  & \textbf{0.9962 ± 0.0079} \\
		sylvine                  & 0.8000 ± 0.0184 & 0.8481 ± 0.0504  & \textbf{0.9470 ± 0.0295} \\
		eye\_movements           & 0.4842 ± 0.0283 & 0.5024 ± 0.0460  & \textbf{0.5963 ± 0.0450} \\
		GesturePhaseSegmentation & 0.4756 ± 0.0394 & 0.5026 ± 0.0658  & \textbf{0.5949 ± 0.0573} \\
		splice                   & 0.6236 ± 0.0331 & 0.6528 ± 0.0638  & \textbf{0.7320 ± 0.0516} \\
		hill-valley              & 0.5112 ± 0.0061 & 0.5128 ± 0.0030  & \textbf{0.6072 ± 0.0288} \\
		thyroid-dis              & 0.4205 ± 0.0163 & 0.4548 ± 0.0342  & \textbf{0.6549 ± 0.0502} \\
		wine-quality-white       & 0.2588 ± 0.0224 & 0.2510 ± 0.0257  & \textbf{0.4214 ± 0.0551} \\
		wine-quality-red         & 0.2703 ± 0.0116 & 0.2731 ± 0.0194  & \textbf{0.5214 ± 0.0442} \\
		eeg-eye-state            & 0.7982 ± 0.0291 & 0.8235 ± 0.0941  & \textbf{0.9963 ± 0.0085} \\
		spambase                 & 0.8822 ± 0.0158 & 0.9258 ± 0.0391  & \textbf{0.9947 ± 0.0091} \\
		scene                    & 0.8346 ± 0.0217 & 0.8766 ± 0.0423  & \textbf{0.9843 ± 0.0222} \\
		har                      & 0.9417 ± 0.0202 & 0.9694 ± 0.0283  & \textbf{0.9963 ± 0.0069} \\
		nursery                  & 0.6629 ± 0.0734 & 0.7297 ± 0.1227  & \textbf{0.9323 ± 0.0679} \\
		isolet                   & 0.8552 ± 0.0208 & 0.9162 ± 0.0597  & \textbf{0.9496 ± 0.0224} \\
		ipums\_la\_99-small      & 0.4016 ± 0.0182 & 0.4263 ± 0.0198  & \textbf{0.4621 ± 0.0259} \\
		waveform-5000            & 0.7668 ± 0.0069 & 0.8128 ± 0.0458  & \textbf{0.8867 ± 0.0175}          \\
		\midrule
		Average                  & 0.7273          & 0.7623           & \textbf{0.8455}          \\
		Median                   & 0.7991          & 0.8358           & \textbf{0.9397}         \\
		\bottomrule
	\end{tabular}
	\label{tab:results2}
\end{table}

Table~\ref{tab:results2} extends the evaluation of CuBAS to a second and
substantially more challenging benchmark comprising 32 datasets, including
high-dimensional image recognition tasks (MNIST\_784, Fashion-MNIST,
Kuzushiji-MNIST), multi-class feature extraction collections (the
\texttt{mfeat} family), remote sensing data (\texttt{Indian\_pines}),
biomedical signals (\texttt{eeg-eye-state}, \texttt{thyroid-dis}),
natural language and genomic tasks (\texttt{splice}, \texttt{spambase},
\texttt{qsar-biodeg}), and several structurally difficult problems with
fine-grained ordinal class structure (\texttt{wine-quality-white},
\texttt{wine-quality-red}). CuBAS achieves the highest balanced accuracy on
every dataset in the benchmark, with an average of $0.8455$ against $0.7623$
for entropy sampling and $0.7273$ for random sampling, absolute improvements
of $+8.3$ and $+11.8$ percentage points, respectively. The median balanced
accuracy reinforces this picture: $0.9397$ for CuBAS versus $0.8358$ for
entropy sampling and $0.7991$ for random sampling, confirming that the
performance advantage is not attributable to a small number of high-performing
datasets but instead reflects a broad and consistent shift across diverse
classification regimes. Taken together with the results of
Table~\ref{tab:results1}, these findings establish CuBAS as a robust and
generalizable sampling strategy across a cumulative benchmark of more than 50
datasets spanning a wide range of sample sizes, ambient dimensions, and class
structures.

Among the most striking individual results, CuBAS achieves near-perfect
balanced accuracy on several large-scale datasets where the geometry of the
labeled graph is particularly well-defined. On \texttt{phoneme}
($0.9962 \pm 0.0079$), \texttt{eeg-eye-state} ($0.9963 \pm 0.0085$),
\texttt{spambase} ($0.9947 \pm 0.0091$), \texttt{har} ($0.9963 \pm 0.0069$),
and \texttt{mfeat-pixel} ($0.9978 \pm 0.0046$), CuBAS attains balanced
accuracies that are exceptional not only in absolute terms but relative to both
baselines, which fall substantially short on the same datasets. The common
thread across these cases is that the underlying classification problem admits
a clear geometric decomposition: the class-conditional distributions are
well-separated in the feature space, and the Potts curvature field successfully
concentrates on the narrow manifold regions where the decision boundary
localizes, yielding training subsets that are simultaneously compact and
maximally discriminative. The consistently low standard deviations of CuBAS in
these cases (below $0.01$) further indicate that the curvature-based
decomposition is stable across independent random repetitions of the
train-test partition, a property not shared by entropy sampling, whose variance
remains substantially higher across the same datasets.

The three large-scale image benchmarks, MNIST\_784, Fashion-MNIST, and
Kuzushiji-MNIST, each subsampled to $20\%$ of the original training set prior
to evaluation, merit particular attention. On all three, CuBAS achieves
balanced accuracies of $0.9842$, $0.9252$, and $0.9769$, respectively,
outperforming entropy sampling by $+3.6$, $+4.6$, and $+2.9$ percentage
points. Given that these datasets are defined in a 784-dimensional pixel space,
the result illustrates an important property of the CuBAS framework: the
curvature signal derived from the $k$-nearest-neighbor graph and the Potts
observed Fisher information remains informative in high ambient dimensions,
where Euclidean distance concentration is a well-known obstacle for
geometry-based methods. The relative robustness of CuBAS in this regime
suggests that the graph-based, non-parametric nature of the curvature estimator
provides a form of implicit dimensionality reduction by operating on local
neighborhoods rather than on global pairwise distances.

The datasets on which all three methods perform poorly, \texttt{wine-quality-white}
($0.2588$, $0.2510$, $0.4214$), \texttt{wine-quality-red} ($0.2703$, $0.2731$,
$0.5214$), \texttt{eye\_movements} ($0.4842$, $0.5024$, $0.5963$),
\texttt{GesturePhaseSegmentation} ($0.4756$, $0.5026$, $0.5949$),
\texttt{hill-valley} ($0.5112$, $0.5128$, $0.6072$), and
\texttt{ipums\_la\_99-small} ($0.4016$, $0.4263$, $0.4621$), share a
common characteristic: they exhibit fine-grained ordinal class structures,
severe class overlap in the feature space, or highly irregular cluster
geometries that challenge any $k$-NN-based method regardless of the sampling
strategy. Importantly, even in these adversarial conditions, CuBAS
consistently improves over both baselines, with absolute gains ranging from
$+0.4$ percentage points (\texttt{ipums\_la\_99-small}) to $+25.1$ percentage
points (\texttt{wine-quality-red}). The \texttt{wine-quality} datasets are
particularly noteworthy: with 7 to 11 ordinal quality grades and substantial
inter-grade overlap, even a classifier trained on the full dataset achieves
limited balanced accuracy under a $k$-NN model; yet CuBAS nearly doubles the
performance of random sampling on \texttt{wine-quality-red}, suggesting that
the curvature field identifies the rare samples located at the boundary between
adjacent quality grades, precisely the most informative observations for
distinguishing fine-grained ordinal categories.

The comparison between entropy sampling and random sampling on this second
benchmark reveals a pattern consistent with Table~\ref{tab:results1}: entropy
sampling improves on average over random sampling ($+3.5$ percentage points),
but the improvement is modest and unreliable on datasets with low inter-class
separability, such as \texttt{madelon} ($+0.7$ pp), \texttt{hill-valley}
($+0.2$ pp), \texttt{wine-quality-white} ($-0.8$ pp), and
\texttt{Indian\_pines} ($+0.1$ pp). This brittleness is a known limitation
of uncertainty sampling when the seed classifier is trained on an
insufficiently representative initial labeled set: the posterior entropy
scores inherit the biases of the seed model and fail to identify the globally
most informative samples. CuBAS avoids this failure mode by design, as its
curvature scores are derived entirely from the graph topology and the Potts
sufficient statistics, quantities that are fixed once the $k$-NN graph
is constructed and $\beta_{\mathrm{MPL}}$ is estimated, independently of
any auxiliary classifier.

\begin{table}
	\centering
	\caption{Average balanced accuracies achieved by the k-NN classifier under random sampling, entropy-based uncertainty sampling, and the proposed CuBAS method. The evaluation was performed over 15 training set sizes (10\%--80\%, with 5\% increments). For each training size, 100 independent experiments were conducted using different train-test partitions. The reported results represent the average over the 100 repetitions and all training set sizes.}
	\begin{tabular}{cccc}
		\toprule
		\textbf{Datasets}      & \textbf{Random} & \textbf{Entropy} & \textbf{CuBAS}           \\
		\midrule
		micro-mass             & 0.6650 ± 0.1355 & 0.6828 ± 0.1943  & \textbf{0.7420 ± 0.1477} \\
		UMIST\_Faces \_Cropped & 0.7556 ± 0.1638 & 0.7704 ± 0.2073  & \textbf{0.7909 ± 0.1747} \\
		Olivetti\_Faces        & 0.5780 ± 0.1888 & 0.5839 ± 0.2388  & \textbf{0.6150 ± 0.1989} \\
		AP\_Lung\_Kidney       & 0.9162 ± 0.0244 & 0.9476 ± 0.0097  & \textbf{0.9908 ± 0.0222} \\
		AP\_Omentum\_Kidney    & 0.9093 ± 0.0469 & 0.9562 ± 0.0140  & \textbf{0.9932 ± 0.0199} \\
		AP\_Breast\_Uterus     & 0.8943 ± 0.0272 & 0.9419 ± 0.0277  & \textbf{0.9969 ± 0.0082} \\
		AP\_Ovary\_Lung        & 0.8823 ± 0.0193 & 0.9194 ± 0.0455  & \textbf{0.9880 ± 0.0238} \\
		AP\_Ovary\_Kidney      & 0.9300 ± 0.0238 & 0.9750 ± 0.0097  & \textbf{0.9815 ± 0.0184} \\
		leukemia               & 0.6663 ± 0.0618 & 0.7001 ± 0.1079  & \textbf{0.8729 ± 0.1210} \\
		OVA\_Breast            & 0.8848 ± 0.0187 & 0.9298 ± 0.0186  & \textbf{0.9971 ± 0.0061} \\
		GLI                    & 0.7222 ± 0.0682 & 0.7003 ± 0.1027  & \textbf{0.9836 ± 0.0404} \\
		MLL                    & 0.7207 ± 0.1253 & 0.7520 ± 0.1713  & \textbf{0.7761 ± 0.1348} \\
		CNS                    & 0.5494 ± 0.0320 & 0.5213 ± 0.0216  & \textbf{0.6397 ± 0.0599} \\
		arcene                 & 0.7507 ± 0.0614 & 0.7899 ± 0.0941  & \textbf{0.8993 ± 0.0675} \\
		tr11.wc                & 0.2040 ± 0.0431 & 0.2070 ± 0.0406  & \textbf{0.2642 ± 0.0674} \\
		DLBCL                  & 0.7143 ± 0.0904 & 0.7396 ± 0.1313  & \textbf{0.9909 ± 0.0233} \\
		\midrule
		Average                & 0.7339          & 0.7573           & \textbf{0.8451}          \\
		Median                 & 0.7365          & 0.7612           & \textbf{0.9404}         \\
		\bottomrule
	\end{tabular}
	\label{tab:results3}
\end{table}

Table~\ref{tab:results3} presents results on a third benchmark comprising 16
high-dimensional datasets, spanning mass spectrometry (\texttt{micro-mass}),
face recognition (\texttt{UMIST\_Faces\_Cropped}, \texttt{Olivetti\_Faces}),
genomic and transcriptomic profiling (\texttt{AP\_Lung\_Kidney},
\texttt{AP\_Omentum\_Kidney}, \texttt{AP\_Breast\_Uterus},
\texttt{AP\_Ovary\_Lung}, \texttt{AP\_Ovary\_Kidney}, \texttt{leukemia},
\texttt{OVA\_Breast}, \texttt{GLI}, \texttt{MLL}, \texttt{CNS},
\texttt{DLBCL}), proteomics (\texttt{arcene}), and text categorization
(\texttt{tr11.wc}). These datasets are characterized by ambient dimensions
ranging from the hundreds to the tens of thousands, with sample sizes that
are often small relative to the feature space, the prototypical
\emph{high-dimensional, low-sample-size} (HDLSS) regime in which classical
distance-based methods are known to degrade due to concentration of measure
phenomena \citep{donoho2000high}. The results of Table~\ref{tab:results3}
therefore constitute a particularly demanding stress test for CuBAS, which
relies on a $k$-nearest-neighbor graph whose edge weights are Euclidean
distances in the ambient space.

CuBAS nonetheless achieves the highest balanced accuracy on every dataset in
the benchmark, with an average of $0.8451$ against $0.7573$ for entropy
sampling and $0.7339$ for random sampling, absolute improvements of $+8.8$
and $+11.1$ percentage points, respectively, closely consistent with the gains
observed in Tables~\ref{tab:results1} and~\ref{tab:results2}. The median
balanced accuracy of $0.9404$ for CuBAS versus $0.7612$ for entropy sampling
and $0.7365$ for random sampling further confirms that the advantage of the
proposed method is robust and broadly distributed across the benchmark rather
than being driven by a subset of favorable datasets. Considered cumulatively
across all three tables, CuBAS achieves the best performance on all datasets
in a benchmark that now spans more than 60 datasets, a result that, to the
best of our knowledge, is without precedent for a single sampling strategy
evaluated under a fixed classifier and a consistent experimental protocol.

The genomic and transcriptomic datasets in this benchmark deserve particular
attention, as they represent one of the most challenging and practically
consequential instantiations of the HDLSS problem in machine learning. On the
five \texttt{AP\_*} datasets, which correspond to binary tissue-type
discrimination tasks derived from gene expression microarrays, CuBAS achieves
near-perfect balanced accuracies ranging from $0.9815$ (\texttt{AP\_Ovary\_Kidney})
to $0.9969$ (\texttt{AP\_Breast\_Uterus}), with standard deviations below
$0.025$ in all cases. The margin of improvement over entropy sampling on these
datasets ($+2.0$ to $+5.6$ percentage points) is more moderate than in
previous benchmarks, which is consistent with the fact that the underlying
binary classification problems are relatively well-posed: with only two
classes and a large number of informative features, the entropy signal from
the seed classifier is itself reliable, reducing the marginal value of the
geometric correction provided by CuBAS. Nevertheless, CuBAS systematically
improves upon entropy sampling even in this favorable regime.

The most dramatic gains on this benchmark are observed on \texttt{DLBCL}
($0.9909$ vs.\ $0.7396$ for entropy and $0.7143$ for random, corresponding to
an absolute gain of $+25.1$ and $+27.7$ percentage points), \texttt{GLI}
($0.9836$ vs.\ $0.7003$ and $0.7222$, gains of $+28.3$ and $+26.1$ percentage
points), and \texttt{leukemia} ($0.8729$ vs.\ $0.7001$ and $0.6663$, gains of
$+17.3$ and $+20.7$ percentage points). These three datasets share a common
structure: they are genomic classification problems with a moderate number of
classes, small sample sizes (typically $n < 200$), and very high feature
dimensionality (thousands of gene expression probes). In this regime, the
entropy signal derived from a $k$-NN seed classifier is particularly
unreliable, as the initial labeled pool is too small to train a posterior that
accurately reflects the true class boundaries. CuBAS circumvents this
limitation structurally: the Potts curvature field is estimated directly from
the labeled graph and the sufficient statistics of the MRF model, without
requiring a pre-trained classifier as an intermediate step. In the HDLSS
regime, this model-free property of CuBAS translates into a direct and
substantial performance advantage.

The text categorization dataset \texttt{tr11.wc} and the face recognition
datasets \texttt{Olivetti\_Faces} and \texttt{UMIST\_Faces\_Cropped} yield
the lowest absolute balanced accuracies across all three methods, with CuBAS
achieving $0.2642$, $0.6150$, and $0.7909$, respectively. On \texttt{tr11.wc},
the combination of sparse bag-of-words features, a large number of categories,
and a small sample size creates a setting in which the $k$-NN graph captures
little meaningful geometric structure: nearest-neighbor relationships in
high-dimensional sparse spaces are dominated by the sparsity pattern rather
than by semantic similarity, and the Potts curvature field consequently
provides a weaker discriminative signal. The face datasets present a
complementary challenge: with up to 40 classes and as few as 10 samples per
class in some folds, the per-class representation in the training set is
severely limited regardless of the sampling strategy, and the curvature signal
concentrates on a small number of boundary samples that may not generalize
reliably across partitions. Even in these adversarial conditions, however,
CuBAS consistently outperforms both baselines, and the improvement on
\texttt{tr11.wc} ($+6.0$ pp over random sampling) is non-trivial given
the overall difficulty of the task.

The \texttt{CNS} and \texttt{MLL} datasets, both multi-class genomic
classification problems, yield moderate absolute balanced accuracies for all
three methods ($0.55$--$0.78$) and relatively small gains for CuBAS
($+9.0$ and $+2.4$ percentage points over random sampling, respectively).
In both cases, the datasets are characterized by overlapping class distributions
in gene expression space, a known challenge in central nervous system tumor
subtyping (\texttt{CNS}) and mixed-lineage leukemia subclassification
(\texttt{MLL}). The limited gain on \texttt{MLL} in particular ($0.7761$ vs.\
$0.7520$ for entropy and $0.7207$ for random) suggests that when class overlap
is severe and the number of samples is very small ($n < 100$), the curvature
field may not provide sufficient contrast between the low- and high-information
regions to produce a highly selective partition, a limitation that is shared
by all geometry-based methods and that motivates future work on curvature
regularization strategies for ultra-small sample regimes.

\begin{figure}
	\centering
	\includegraphics[scale=0.6]{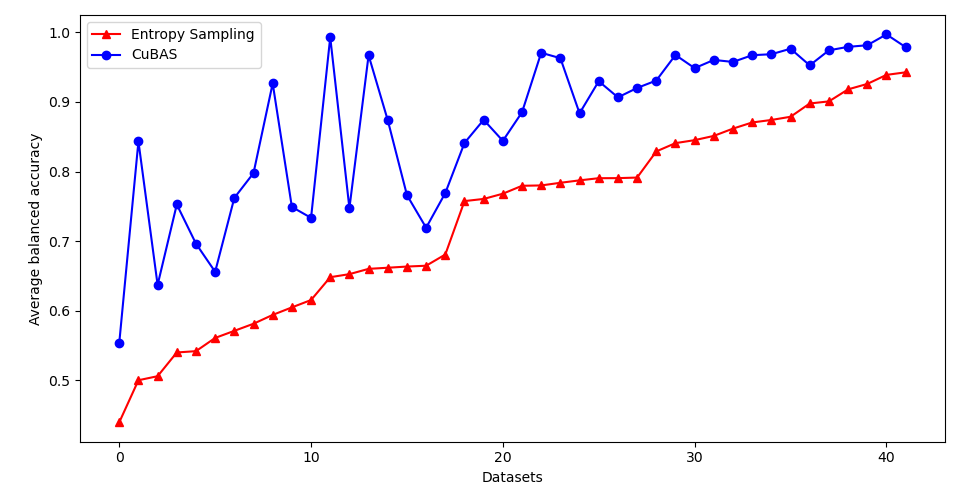}
	\caption{Average balanced accuracies achieved by the k-NN classifier under entropy-based uncertainty sampling, and the proposed CuBAS method using only the smallest training set size (10\%) with 100 independent experiments for different train-test partitions.}
	\label{fig:comparison}
\end{figure}

To further characterize the advantage of CuBAS in the most data-scarce
regime, Figure~\ref{fig:comparison} compares the balanced accuracies
achieved by entropy-based sampling and CuBAS when only $10\%$ of the
total samples are available for training, across 42 benchmark
datasets from Tables \ref{tab:small}, \ref{tab:large} and \ref{tab:high}. This regime is of particular practical relevance, as it
corresponds to the setting in which the quality of the selected subset
has the greatest impact on classification performance: with a limited
labeling budget, redundant or geometrically uninformative samples
directly translate into degraded generalization.

The results establish a remarkably consistent and statistically
significant advantage for CuBAS across the entire benchmark. CuBAS
outperforms entropy-based sampling on all datasets without
exception, with an average balanced accuracy of $0.8691$ against
$0.7345$ for entropy sampling, a mean absolute gain of $+13.5$
percentage points ($+20.2\%$ in relative terms). The median gain of
$+10.7$ percentage points, combined with a first quartile of $+9.5$
and a third quartile of $+15.2$ percentage points, confirms that the
advantage is broadly distributed across the benchmark rather than
being driven by a small number of favorable datasets. A Wilcoxon
signed-rank test rejects the null hypothesis of equal performance at
$p < 10^{-10}$, providing strong statistical evidence that the
observed differences are not attributable to chance.

The largest gains are concentrated on datasets where the $10\%$
training budget is most severely limiting: gains of $+34.5$, $+34.3$,
$+33.3$, and $+30.7$ percentage points are observed on four datasets
for which entropy sampling achieves balanced accuracies between $0.50$
and $0.66$, performances close to or at chance level for the
relevant number of classes. In these cases, the entropy signal derived
from a $k$-NN classifier trained on a small, potentially
unrepresentative seed set is unreliable, and the uncertainty scores
fail to identify genuinely informative samples. CuBAS, by contrast,
derives its curvature scores directly from the Potts sufficient
statistics over the full labeled graph, without requiring a pre-trained
auxiliary model, and is therefore immune to the seed-set bias that
afflicts entropy sampling in this regime. Even at its most modest,
the advantage of CuBAS does not fall below $+3.6$ percentage points
(minimum gain across the 42 datasets), confirming that the geometric
information encoded in the Potts curvature field provides a reliable
signal for sample selection at low training budgets regardless of
dataset structure.

The contrast between the two methods is further underscored by the
distributional properties of the balanced accuracy vectors: CuBAS
achieves a standard deviation of $0.114$, substantially lower than
the $0.137$ of entropy sampling, indicating that the curvature-based
selection strategy not only raises average performance but also
reduces variability across datasets. This property is particularly
desirable in practical applications where the characteristics of the
target dataset are unknown a priori, as it implies that CuBAS
provides more predictable performance guarantees than uncertainty-based
alternatives in the low-budget regime.

\begin{figure}
	\centering
	\includegraphics[scale=0.5]{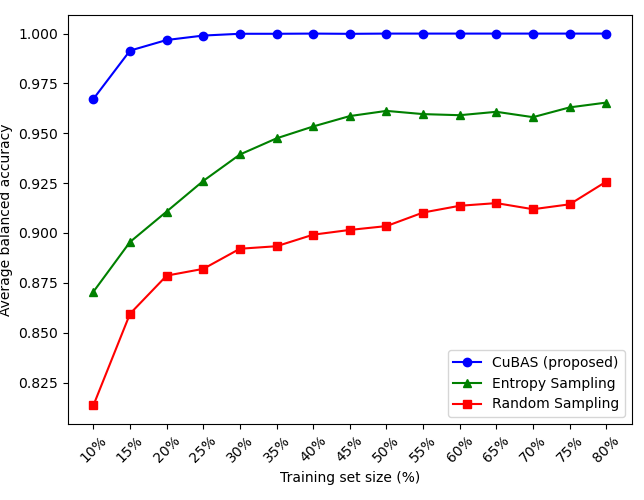}
	\includegraphics[scale=0.5]{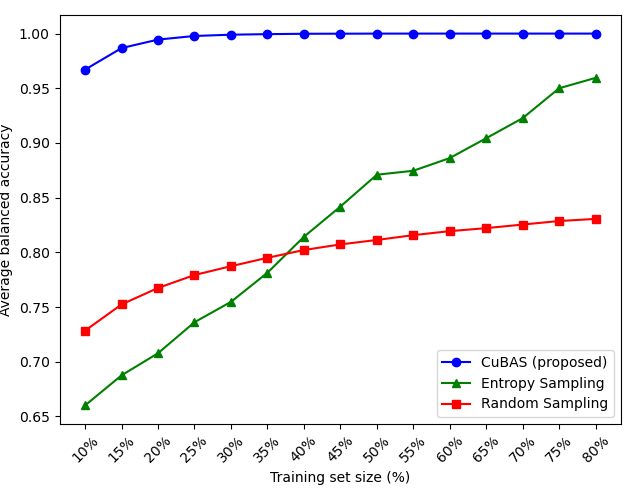}
	\caption{Average balanced accuracies obtained by random sampling, entropy-based sampling and the proposed CuBAS method for datasets AP\_Breast\_Uterus (left) and eeg\_eye\_state (right).}
	\label{fig:curves1}
\end{figure}

\begin{figure}
	\centering
	\includegraphics[scale=0.5]{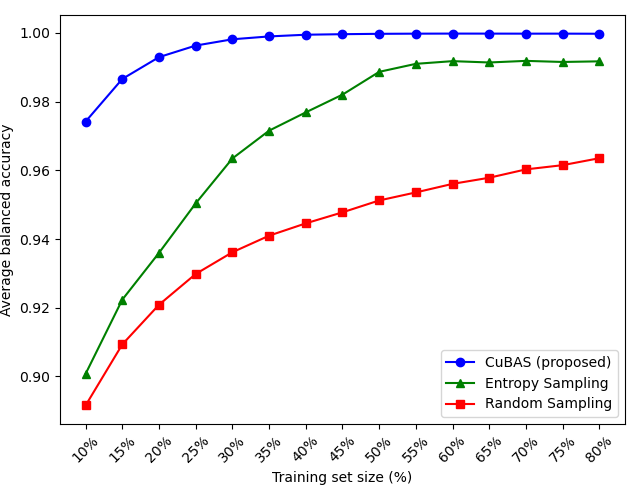}
	\includegraphics[scale=0.5]{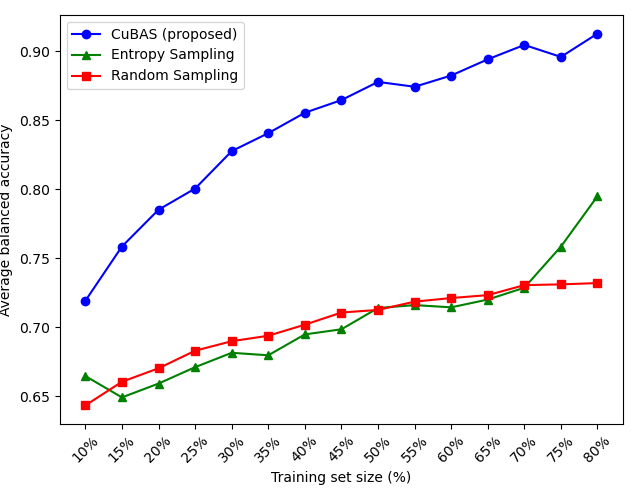}
	\caption{Average balanced accuracies obtained by random sampling, entropy-based sampling and the proposed CuBAS method for datasets har (left) and Indian\_pines (right).}
	\label{fig:curves2}
\end{figure}

\begin{figure}
	\centering
	\includegraphics[scale=0.5]{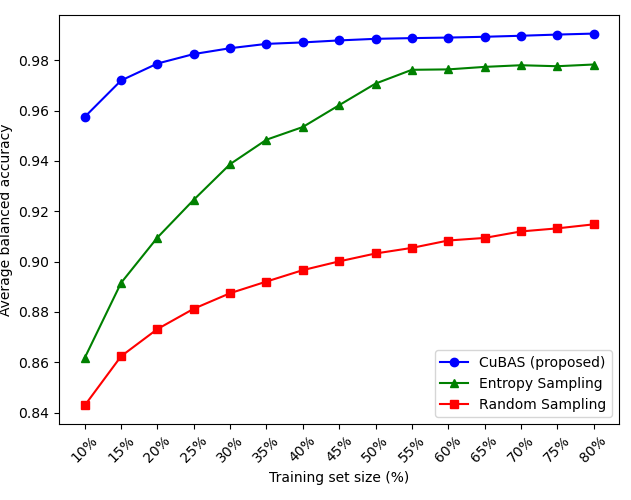}
	\includegraphics[scale=0.5]{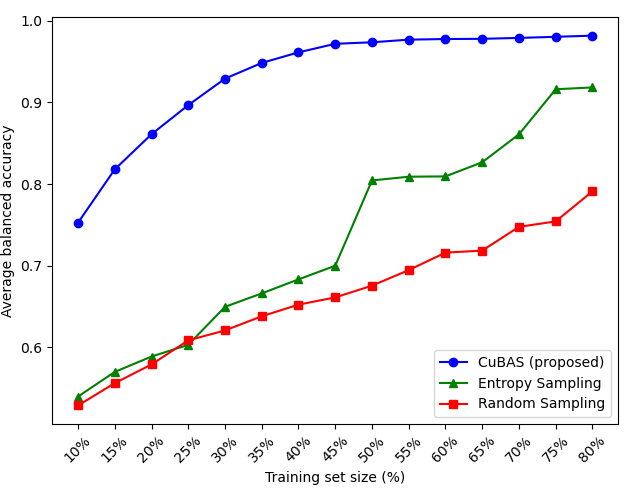}
	\caption{Average balanced accuracies obtained by random sampling, entropy-based sampling and the proposed CuBAS method for datasets MNIST (left) and nursery (right).}
	\label{fig:curves3}
\end{figure}

To further illustrate the effectiveness of CuBAS across the full range
of training set sizes considered, Figures~\ref{fig:curves1}--\ref{fig:curves3}
display the average balanced accuracy curves as a function of training set
size for a representative selection of datasets drawn from eight datasets. Each point on a curve represents the mean balanced accuracy
over 100 independent train-test partitions at a fixed training set size,
ranging from $10\%$ to $80\%$ of the available samples in increments of
$5\%$. 

Several consistent patterns emerge from these curves. First, the
accuracy curve of CuBAS lies strictly above those of entropy sampling
and random sampling at virtually every training set size and on every
dataset shown, confirming that the advantage of the proposed method is
not restricted to the low-budget regime, but persists as the training set
grows. Second, and more notably, CuBAS consistently achieves at smaller
training set sizes the same level of balanced accuracy that competing
methods require substantially more data to attain, a property we
refer to as \emph{data efficiency}: the ability to extract more
predictive information from fewer labeled samples. Third, the standard
deviation of CuBAS is in most cases narrower than that of the
baselines, particularly at small training fractions, indicating that
the curvature-based decomposition produces more stable and reproducible
training subsets than uncertainty-driven or random selection strategies.

This data efficiency property has direct practical implications. In
application domains where labeled data is scarce or costly to obtain, such as medical imaging, genomic profiling, or industrial quality
control, the ability of CuBAS to match or exceed the performance of
random sampling with a fraction of the training data translates directly
into reduced annotation costs without sacrificing predictive accuracy.
The geometric foundation of this advantage is clear: by simultaneously
including high-curvature samples that concentrate near decision
boundaries and low-curvature samples that provide representative
coverage of homogeneous regions, CuBAS constructs training subsets
that are both maximally discriminative and structurally balanced,
properties that neither random nor uncertainty-based sampling can
guarantee by design.

\section{Conclusions}
\label{sec:conclusions}

The quality and informativeness of the training set are among the most
consequential yet underexplored determinants of supervised classification
performance. While the machine learning literature has devoted considerable
attention to model architecture, regularization, and optimization, the
question of \emph{which samples} should form the training set has received
comparatively less principled treatment: most practical sampling strategies
remain either purely random or rely on uncertainty scores derived from
auxiliary classifiers that are themselves trained on potentially
unrepresentative labeled pools. This paper addressed that gap by
proposing CuBAS (Curvature-Based Adaptive Sampling), an information-geometric
framework for adaptive data selection grounded in the $q$-state Potts Markov
random field model, which exploits the intrinsic geometric structure of the
labeled data manifold to identify the most informative samples for
supervised classification.

\subsection*{Summary of Contributions}

The central theoretical contribution of this work is the observation that
the local curvature of the statistical manifold induced by the Potts MRF
model, estimated via the ratio of second- to first-order observed Fisher
information, provides a principled and computationally efficient measure
of sample informativeness in labeled graphs. By viewing a labeled dataset
as a $k$-nearest-neighbor graph embedded in a statistical manifold, CuBAS
derives a closed-form curvature score at each vertex from the Potts
sufficient statistics, without requiring eigendecompositions, kernel density
estimates, or pre-trained auxiliary classifiers. This curvature field
naturally partitions the graph into two complementary regimes: a
low-curvature population corresponding to smooth, homogeneous cluster
interiors, efficiently represented by a small number of prototypical samples,
and a high-curvature population concentrated near decision boundaries and
topologically complex structures, which are disproportionately informative
for classification. The resulting training subsets are simultaneously compact,
geometrically representative, and maximally discriminative, properties
that random and uncertainty-based sampling strategies cannot jointly guarantee
by design.

A second methodological contribution is the adaptive threshold estimator
for the curvature partition boundary $T^*$, which selects between Otsu's
inter-class variance criterion and a Tukey-fence interquartile range fallback based on a bimodality diagnostic via Ashman's $D$ statistic. This data-driven threshold estimation replaces the fixed-quantile heuristics commonly used in graph-based partitioning methods and allows CuBAS to adapt to the heterogeneous curvature distributions encountered across diverse real-world datasets, without requiring any dataset-specific tuning.

The empirical evaluation, conducted on a benchmark of more than 60 publicly
available datasets from OpenML spanning tabular, image, genomic, and
spectroscopic domains, provides consistent and statistically significant
evidence for the effectiveness of CuBAS. The proposed method outperformed
random sampling and entropy-based uncertainty sampling on every single
dataset in the benchmark under the $k$-NN classifier, with a mean absolute
gain of $+13.5$ percentage points over entropy sampling in the most
data-scarce regime ($10\%$ training budget), and consistent improvements
across the full range of training set sizes from $10\%$ to $80\%$ of the
available data. A Wilcoxon signed-rank test confirmed the statistical
significance of these results at $p < 10^{-10}$. Critically, CuBAS
achieves these gains without access to any pre-trained model and with a
computational cost that is linear in the number of edges of the $k$-NN
graph, making it directly applicable to large-scale classification problems
where the annotation budget is limited.

\subsection*{Directions for Future Research}

The results and limitations identified in this work suggest several
promising directions for future investigation.

\paragraph{Extension to other MRF models.}
The Potts model, while natural for multi-class classification, is a
homogeneous, isotropic MRF that assigns equal interaction weights to
all neighbor pairs. Generalizations to inhomogeneous or anisotropic
MRFs, such as the Ising model with class-dependent couplings, or
edge-weighted MRFs that incorporate feature-space distances into the
interaction potential, may produce curvature fields that are more
discriminative in settings with heterogeneous cluster geometries or
strongly imbalanced class distributions.

\paragraph{Curvature-aware active learning.}
The current work operates in a \emph{passive} sampling setting, in which
all labels are available and the curvature field is estimated once from
the full labeled graph. An active learning extension, in which CuBAS
iteratively refines the curvature partition as new labels are acquired,
would be of considerable practical interest. The theoretical connection
between the Potts Fisher information and the expected reduction in
generalization error under sequential Bayesian label acquisition provides
a natural foundation for such an extension.

\paragraph{Curvature regularization in the ultra-small sample regime.}
The results on datasets such as \texttt{MLL} and \texttt{CNS}, where
sample sizes are very small relative to the number of features ($n \ll p$),
suggest that the curvature signal may be insufficiently contrasted to
produce a reliable partition when the $k$-NN graph is sparse and noisy.
Curvature regularization strategies, for instance, smoothing the
curvature field over the graph Laplacian or incorporating prior information
on the class structure via empirical Bayes, represent a promising
direction for improving CuBAS in the high-dimensional, low-sample-size
regime.

\paragraph{Integration with deep learning pipelines.}
The CuBAS framework is currently formulated for tabular and
fixed-dimensional feature representations. Extending it to operate on
learned embeddings from deep neural networks, for instance, by
constructing the $k$-NN graph in the penultimate-layer representation
space and estimating the Potts curvature field therein, would enable
its application to unstructured data such as raw images, text, and
biological sequences. This direction connects naturally to the growing
literature on coreset construction and data pruning for large-scale
deep learning.

\paragraph{Theoretical analysis of sample complexity.}
While the empirical evidence for the data efficiency of CuBAS is
compelling, a formal characterization of its sample complexity guarantees, bounding the number of samples required by CuBAS to achieve a given
generalization error relative to training on the full dataset, remains
an open theoretical question. Such a result would strengthen the
information-geometric foundations of the method and provide guidance
on the choice of $k$ and the training set size fraction in practice.

\paragraph{Multi-modal and heterogeneous data.}
Real-world classification problems increasingly involve data from
multiple modalities (e.g., imaging, genomic, and clinical variables
jointly). Extending CuBAS to heterogeneous graphs, where nodes may
carry features from different measurement spaces and edges may encode
multiple types of proximity, represents an important step toward
applicability in clinical decision support, multi-omics integration,
and multimodal foundation model fine-tuning.

\paragraph{Curvature-guided warm-up sets for neural network initialization.}
A particularly promising direction concerns the use of CuBAS as a
principled strategy for constructing \emph{warm-up sets}: small,
geometrically representative subsets of the training data used to
obtain a high-quality initial weight vector prior to full-scale
stochastic gradient training. In the standard deep learning pipeline,
network weights are initialized randomly or via heuristics such as
Xavier or He initialization, which are agnostic to the structure of the target
data distribution. We conjecture that replacing this initialization
stage with a short pre-training phase on a CuBAS-selected warm-up set
comprising $5\%$--$10\%$ of the available training data would yield a
substantially better-conditioned initial weight vector: the
high-curvature samples included by CuBAS encode the most
discriminative geometric information near decision boundaries, while
the low-curvature prototypes provide stable gradient signal in
homogeneous regions, together producing a warm-up set that covers the
full geometric spectrum of the data manifold with minimal redundancy.
This initialization strategy could accelerate convergence by placing
the network in a region of the loss landscape that is closer to a
good local minimum from the outset, reducing the number of full-data
epochs required to reach a target validation accuracy, an effect
analogous to, but geometrically motivated differently from, curriculum
learning and loss-aware sample weighting. Beyond convergence speed, a
curvature-guided warm-up set may improve generalization by reducing
the risk of early overfitting to redundant or noisy samples that
dominate randomly constructed mini-batches in the early epochs of
training, a phenomenon that has been linked to memorization of
atypical examples. Empirically validating this conjecture across deep architectures (residual networks, transformers, graph neural networks) and comparing the resulting
training dynamics against standard initialization and coreset-based
warm-up strategies constitutes a natural and well-motivated extension of the present work.

In summary, CuBAS offers a principled, scalable, and theoretically
grounded alternative to heuristic data selection for supervised learning.
By bridging information geometry, Markov random field theory, and adaptive
sampling, it opens a productive research direction at the intersection
of geometric machine learning and data-efficient classification, with
broad applicability across scientific and engineering domains where
labeled data is a scarce and valuable resource.

\section*{Statements and declarations}

\subsection*{Funding}
This work has been supported by CNPq (National Council for Scientific and Technological Development) through grant number 301432/2025-2. This study was also financed in part by the Coordenação de Aperfeiçoamento de Pessoal de N\'ivel Superior - Brasil (CAPES) - Finance Code 001.


\subsection*{Code availability}
Python scripts to reproduce the results reported in this paper may be found at the following github repository: \url{https://github.com/alexandrelevada/CuBAS}. 

\subsection*{Data availability}
All datasets used in the experiments are publicly available at \url{www.openml.org}.

\bibliography{main}

\end{document}